\documentclass{article} % For LaTeX2e
\usepackage[preprint]{colm2026_conference}

\usepackage{microtype}
\usepackage{hyperref}
\usepackage{url}
\usepackage{booktabs}
\usepackage{graphicx}
\usepackage{amsmath}
\usepackage{amssymb}
\usepackage{xcolor}
\usepackage{lineno}
\usepackage{multirow}
\usepackage{tabularx}
\usepackage{placeins}
\newcolumntype{Y}{>{\raggedright\arraybackslash\hspace{0pt}}X}

\definecolor{darkblue}{rgb}{0, 0, 0.5}
\definecolor{distillgray}{gray}{0.62}
\definecolor{rlvrgray}{gray}{0.38}
\hypersetup{
    colorlinks=true,
    citecolor=darkblue,
    linkcolor=darkblue,
    urlcolor=darkblue,
    pdftitle={Dense Supervision, Sparse Updates: On the Sparsity and Geometry of On-Policy Distillation},
    pdfauthor={Guo Yu, Hao-Xuan Ma, Jun-Peng Jiang, Han-Jia Ye, Wenlin Liu, and Yulan Hu}
}

\title{Dense Supervision, Sparse Updates: \\On the Sparsity and Geometry of On-Policy Distillation}

\author{Guo Yu, Hao-Xuan Ma, Jun-Peng Jiang, Han-Jia Ye\thanks{Corresponding authors.} \\
School of Artificial Intelligence, Nanjing University \\ 
National Key Laboratory for Novel Software Technology, Nanjing University \\
\And
Wenlin Liu, Yulan Hu\footnotemark[1] \\
Amap, Alibaba Group
}

\begin{document}

\ifcolmsubmission
\linenumbers
\fi

\maketitle

\begin{abstract}
On-policy distillation (\textsc{OPD}) has recently become a prominent post-training recipe by combining two desirable ingredients: on-policy student-generated trajectories and dense token-level teacher supervision.
Yet how this hybrid training regime shapes a model remains poorly understood.
We characterize the sparsity and geometry of \textsc{OPD} parameter updates across several language and vision-language model pairs and application settings.
\textsc{OPD} updates are small and coordinate-sparse, while remaining distributed across layers and modules.
This sparse support is operationally meaningful: masked training on the discovered subnetwork nearly recovers full-training performance.
At the matrix level, the updates are numerically full-rank but spectrally concentrated.
Their visible supports avoid coordinates emphasized by the source's principal structure and favor low-magnitude source coordinates, while the source singular-value spectra change little.
Together, these findings show that \textsc{OPD} exhibits important weight-space signatures of on-policy post-training despite using dense teacher supervision.
\end{abstract}

\section{Introduction}

On-policy distillation (\textsc{OPD}) is emerging as a third major component in post-training pipelines for large language models, alongside supervised fine-tuning and reinforcement learning \citep{yang2025qwen3,lu2025onpolicydistillation,li2026rethinking,xiaomi2026mimov2flash,deepseekai2026deepseekv4}.
Recent practice uses \textsc{OPD} to transfer frontier-model reasoning into lightweight students, unify multiple post-training capabilities, or turn privileged information into self-improvement signals \citep{song2026surveyopd}.

The reason for this momentum is conceptually intuitive.
Conventional supervised fine-tuning and offline distillation provide dense token-level supervision, but train on fixed demonstrations that may differ from what the model will generate at test time, where distribution shift can cause compounding errors \citep{ross2011reduction,bengio2015scheduled}.
\textsc{RLVR} trains on the current policy's own samples, but learns from environment- or verifier-derived outcome rewards that are often sparse, and therefore faces a credit-assignment problem.
\textsc{OPD}, motivated by on-policy imitation approaches, combines the attractive parts of both: it keeps the data on-policy from the student model while replacing environmental reward feedback with dense teacher supervision \citep{agarwal2024onpolicy}.
Table~\ref{tab:training-regimes} summarizes this conceptual position relative to \textsc{SFT}, offline sequence-level distillation, and \textsc{RLVR}.

\begin{table*}[t]
\centering
\small
\setlength{\tabcolsep}{12pt}
\resizebox{\textwidth}{!}{
\begin{tabular}{@{}llll@{}}
\toprule
Training regime & Data distribution & Learning signal & Analogy \\
\midrule
\textsc{SFT} & Fixed data & Dense token labels & Behavioral Cloning in imitation learning \\
\textsc{SeqKD} & Teacher demonstrations & Dense token labels & Behavioral Cloning in imitation learning \\
\textsc{RLVR} & Current policy & Sparse trajectory reward & Policy-based reinforcement learning \\
\textsc{OPD} & Current policy & Dense teacher feedback & Interactive imitation learning \\
\bottomrule
\end{tabular}
}
\caption{Conceptual position of \textsc{OPD} relative to \textsc{SFT}, offline \textsc{SeqKD}, and \textsc{RLVR}. \textsc{OPD} combines on-policy student samples with dense teacher feedback.}
\label{tab:training-regimes}
\end{table*}

Despite its empirical success, how \textsc{OPD} changes a model remains less understood.
Prior post-training pipelines were often organized as \textsc{SFT} followed by \textsc{RLVR}, and recent work shows that these two stages can behave very differently in weight space.
\citet{mukherjee2025reinforcementlearningfinetunessmall} find that densely supervised off-policy fine-tuning produces much denser parameter updates than sparse-reward \textsc{RLVR}, which instead modifies a relatively small subnetwork.
\citet{zhu2025pathtakenrlvrprovably} further argue that \textsc{RLVR} moves away from the principal directions of the source weights.
These findings leave a gap precisely where \textsc{OPD} sits: if the data are on-policy, but the learning signal comes from dense teacher feedback rather than sparse environment reward, should the update resemble supervised fine-tuning, \textsc{RLVR}, or constitute a distinct regime?
Answering this question tests whether sparse task rewards are necessary for sparse, concentrated adaptation and identifies which parameter structures carry the gains from dense teacher feedback.
This matters for parameter-efficient post-training: coordinate support and singular directions offer two complementary ways to concentrate adaptation in a restricted parameter space.

This paper answers this question by analyzing the sparsity and geometry of \textsc{OPD} parameter updates.
We compute parameter updates between source and trained models, measure their norm and coordinate sparsity, localize their layer and module structure, compare update-support overlap across algorithms, teachers, and data, and analyze their singular spectra, source-spectrum drift, and source-derived coordinate masks.
Results across several model pairs give a coherent picture: \textsc{OPD}-style updates are small in relative norm, sparse at checkpoint precision, distributed across layers and modules, numerically full-rank yet spectrally concentrated, and strongly biased away from source-principal coordinates toward coordinates where the source weights are close to zero, while largely preserving the source singular-value spectrum.
\textsc{OPD} update supports also overlap with \textsc{RLVR} and with \textsc{OPD} supports obtained under different teachers or training data far above independent-support baselines.
In a matched R1-Qwen comparison, a norm-matched reconstruction from the leading 1\% singular components recovers the full \textsc{OPD} checkpoint's reasoning gains, whereas the same intervention fails for \textsc{SFT}.

We further test whether these signatures matter operationally by restarting \textsc{OPD} from the source model and training only coordinates selected by thresholded checkpoint-delta masks.
The mask discovered from the full \textsc{OPD} run recovers nearly the same reasoning performance as full training, while density-matched random masks generally perform worse.

Our contribution is an empirical characterization of \textsc{OPD} update structure across model families and applications, together with functional evidence that selected coordinates and leading update directions retain its reasoning gains.
These findings show that dense teacher supervision is compatible with sparse, spectrally concentrated updates, extending geometric patterns observed in \textsc{RLVR} to \textsc{OPD}.

\section{Background}

\subsection{\textsc{OPD} as on-policy dense supervision}

\textsc{OPD} can be formulated as dense teacher supervision applied to on-policy student behavior \citep{agarwal2024onpolicy}.
In ordinary supervised fine-tuning or offline knowledge distillation, the model is optimized on a fixed dataset of target sequences \citep{hinton2015distilling,kim2016sequence}.
In \textsc{RLVR}, the model samples from its current policy and receives an environment- or verifier-derived reward signal, often at the sequence level \citep{shao2024deepseekmath}.
In \textsc{OPD}, the student also samples from its current policy, but the teacher supplies token-level feedback on the sampled sequence.
This gives \textsc{OPD} the data distribution of on-policy learning and the dense supervision of distillation.

Formally, let \(\pi_\theta\) be the student policy and \(\pi_T\) be the teacher.
For a prompt \(x \sim \mathcal{D}\), \textsc{OPD} first samples a response \(y=(y_1,\ldots,y_{|y|})\sim\pi_\theta(\cdot|x)\), and then minimizes a teacher-student divergence on the student-generated trajectory:
\begin{equation}
    \mathcal{L}_{\mathrm{OPD}}(\theta)
    =
    \mathbb{E}_{\substack{x\sim\mathcal{D}\\y\sim\pi_\theta(\cdot|x)}}
    \left[
    \sum_{t=1}^{|y|}
    D_f\!\left(
        \pi_T(\cdot|x,y_{<t})
        \,\|\, 
        \pi_\theta(\cdot|x,y_{<t})
    \right)
    \right],
\end{equation}
where \(D_f\) can be instantiated as forward / reverse KL divergence, JS divergence, or another token-level $f$-divergence.
The key point is that the conditioning prefixes \(y_{<t}\) come from the student, not from a fixed teacher trace.

This distinction turns \textsc{OPD} into a useful diagnostic for post-training dynamics.
If sparse \textsc{RLVR} updates were mainly a consequence of sparse environment rewards, then replacing those rewards with dense teacher-derived feedback might be expected to produce much denser parameter changes.
If instead the key ingredient is that the model trains near its own policy distribution, then \textsc{OPD} should retain part of the sparse and off-principal structure observed in \textsc{RLVR}, despite changing both the source and granularity of the learning signal.

\subsection{Current uses of \textsc{OPD}}
\label{sec:opd-current-uses}

\textsc{OPD} has been widely adopted in large language model post-training.
Its current uses mainly fall into three groups.

The first is \textit{cross-scale capability transfer}: a stronger, typically larger, external teacher provides token-level targets on trajectories sampled by a smaller student \citep{agarwal2024onpolicy,gu2024minillm}.
Qwen3, for example, applies this pattern to transfer capabilities from larger models to smaller dense and mixture-of-experts students \citep{yang2025qwen3}.
Here, \textsc{OPD} reduces the train--test trajectory mismatch of offline distillation while retaining the deployment benefits of the smaller student.

The second is \textit{multi-capability consolidation}: domain- or stage-specialized policies supervise a single student, often of comparable scale.
MiMo-V2-Flash, DeepSeek-V4, Kimi K3, and MiniCPM5 distill separately trained domain experts into one release model, whereas GLM-5 uses cross-stage \textsc{OPD} to recover capabilities that can regress during sequential post-training \citep{xiaomi2026mimov2flash,ma2026mopd,glm5team2026glm5,deepseekai2026deepseekv4,kimiteam2026kimik3openfrontier,openbmb2026minicpm5}.
These systems differ in teacher selection and stage placement, but in each case \textsc{OPD} serves as behavioral consolidation on the student's own trajectories.

The third is \textit{on-policy self-distillation}: teacher and student roles are played by the same model under asymmetric information.
In OPSD, the teacher conditions on a verified reference trajectory while the student observes only the question, and their predictions are matched on student-generated prefixes \citep{zhao2026selfdistilledreasoner}.
SDPO instead treats the current model conditioned on textual environment feedback---or on a successful rollout used as feedback---as the teacher and distills its feedback-informed predictions into the policy without that feedback \citep{hubotter2026rlselfdistillation}.
These methods retain dense supervision on student trajectories while replacing a separately trained teacher with a privileged-context version of the student.

\subsection{Parameter dynamics in LLM post-training}

Recent work has begun to examine parameter updates alongside reward curves and benchmark scores to understand post-training.
\citet{mukherjee2025reinforcementlearningfinetunessmall} show that \textsc{RLVR} often changes only a small subnetwork of an LLM, with subnetwork-only fine-tuning recovering the full run.
\citet{zhu2025pathtakenrlvrprovably} argue that this apparent sparsity reflects a model-conditioned geometry: \textsc{RLVR} updates are biased away from the principal directions of the source weights and toward low-curvature, spectrum-preserving regions.
These findings raise the question of what parameter characteristics \textsc{OPD} induces.

\begin{figure*}[t]
\centering
\includegraphics[width=\textwidth]{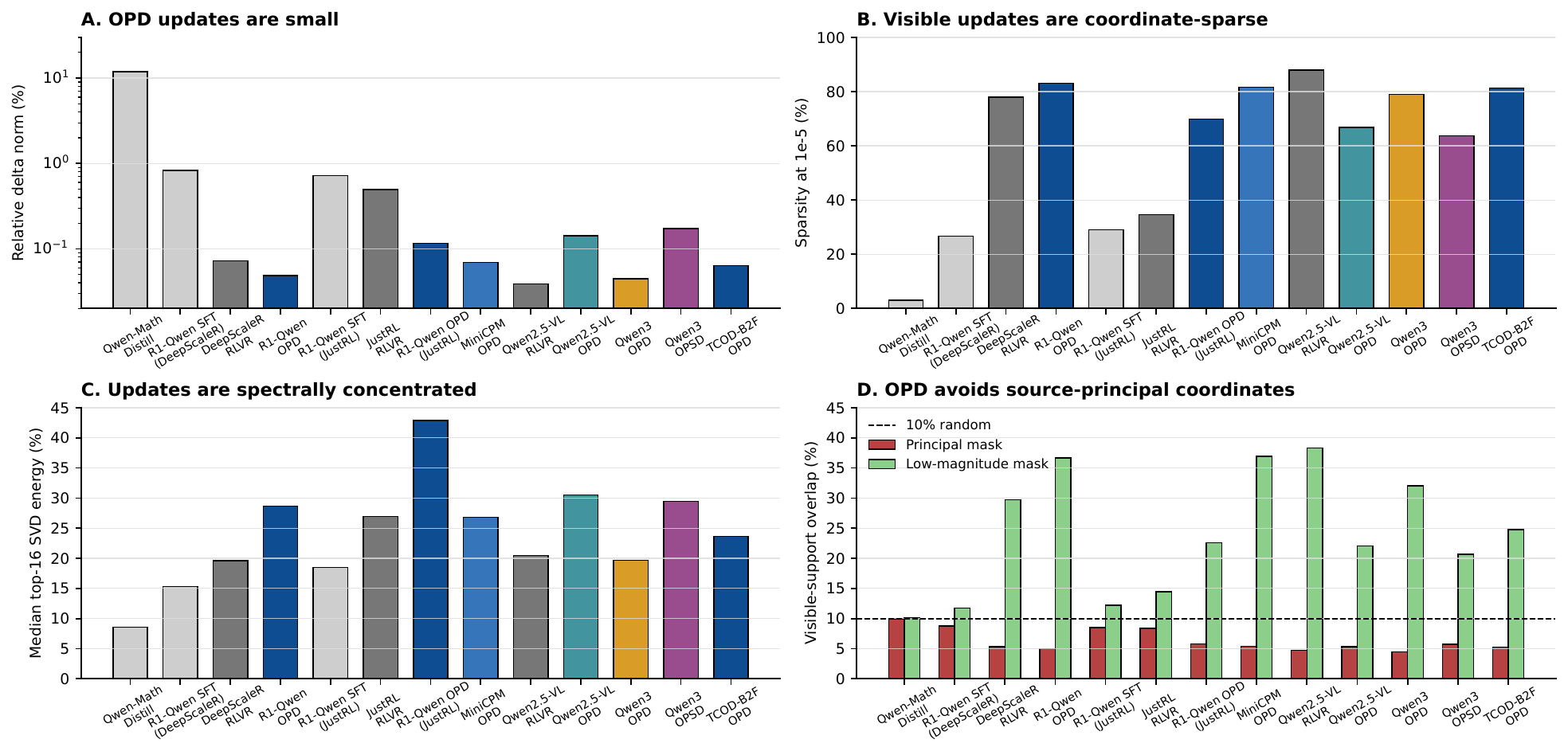}
\caption{Checkpoint deltas show that \textsc{OPD}-style updates are small, coordinate-sparse, spectrally concentrated, and biased away from source-principal coordinates. Gray bars denote offline \textsc{SFT}/\textsc{SeqKD} and \textsc{RLVR} references.}
\label{fig:static-summary}
\end{figure*}

\section{Parameter-update analysis overview}
\label{sec:static-setup}

We use model checkpoint deltas between source and trained models \(\Delta W = W_{\mathrm{trained}} - W_{\mathrm{src}}\) to ask three complementary questions about what \textsc{OPD} changes in weight space.
First, scale and coordinate-support metrics ask whether the final update is globally small and whether it is written into many or few visible coordinates.
Second, spectral metrics ask whether 2D weight-matrix updates are low-rank in a strict sense, or instead full-rank with concentrated singular-value energy.
Third, source-geometry metrics ask whether updates preserve the source singular-value spectrum and whether their visible supports favor source-principal or low-magnitude coordinates.
We compare both endpoints in BF16, a standard release precision for the models studied, and subtract their float32-promoted values; ``at checkpoint precision'' refers to changes visible in this common BF16 representation.
Each metric is introduced where it is first used below; additional implementation details and interpretations are provided in Appendix~\ref{app:delta-metrics}.

We analyze 13 model pairs; Appendix~\ref{app:model-pairs-training} lists their checkpoints, data, and roles.
Seven are \textsc{OPD}-style pairs: R1-Qwen\footnote{DeepSeek-R1-Distill-Qwen-1.5B, an offline-distilled model in the Qwen2.5-Math family.} distilled separately from DeepScaleR and JustRL \citep{luo2025deepscaler,he2025justrl}, MiniCPM5-1B-SFT to MiniCPM5-1B OPD \citep{openbmb2026minicpm5}, Qwen2.5-VL-3B-Instruct distilled from NoisyRollout-7B \citep{bai2025qwen25vl,liu2025noisyrollout}, Qwen3-1.7B-Base distilled from Qwen3-4B-Base-GRPO \citep{li2026rethinking}, Qwen3-4B to an OPSD checkpoint \citep{yang2025qwen3,zhao2026selfdistilledreasoner}, and Qwen2.5-7B-Instruct to TCOD-B2F for agent tasks \citep{wang2026tcod}.
The references comprise three fixed-data \textsc{SFT}/\textsc{SeqKD} pairs and three \textsc{RLVR} pairs spanning DeepScaleR, JustRL, and Qwen2.5-VL \citep{yang2024qwen25math,deepseekai2025deepseekr1}.
An additional R1-Qwen \textsc{OPD} run on OpenThoughts3 provides a data-variation comparison for support overlap and intermediate checkpoints for training-trajectory analysis \citep{guha2025openthoughts}.
Together, the pairs cover LLM and VLM settings, math reasoning and agent tasks, and the three applications in Section~\ref{sec:opd-current-uses}.

\section{Sparsity of \textsc{OPD} updates}
\label{sec:sparsity}

\subsection{Global norm and coordinate sparsity}

For a set \(B\) of matched tensors, we measure the aggregate relative update norm and visible coordinate sparsity as
\begin{equation}
\begin{aligned}
r(B)
&=
\frac{\sqrt{\sum_{W\in B}\|\Delta W\|_F^2}}
     {\sqrt{\sum_{W\in B}\|W_{\mathrm{src}}\|_F^2}},
\\
s_\epsilon(B)
&=
\frac{\sum_{W\in B}|\{i:|\Delta W_i|\leq\epsilon\}|}
     {\sum_{W\in B}|W|}.
\end{aligned}
\end{equation}
Table~\ref{tab:global-delta} uses all matched tensors for \(B\) and \(\epsilon=10^{-5}\), following the visibility convention in prior \textsc{RLVR} checkpoint analysis \citep{mukherjee2025reinforcementlearningfinetunessmall}.

At checkpoint precision, \textsc{OPD}-style deltas are \textbf{consistently small and coordinate-sparse}.
Across all seven pairs, most coordinates remain below the \(10^{-5}\) visibility threshold, whereas the fixed-data \textsc{SFT} references are larger and denser and the offline distillation contrast is larger still and nearly dense (Table~\ref{tab:global-delta}).
In the matched R1-Qwen/DeepScaleR comparison, for example, \textsc{OPD} has \(0.0488\%\) relative norm and \(83.04\%\) visible sparsity, compared with \(0.8280\%\) and \(26.63\%\), respectively, for fixed-data \textsc{SFT}.
The TCOD-B2F agent model similarly has a small relative update norm of \(0.0633\%\) and visible sparsity of \(81.38\%\), extending this pattern to 7B agent training on ALFWorld.
The OPD--SFT sparsity ordering in both R1-Qwen comparisons persists under exact equality, absolute thresholds from \(10^{-6}\) to \(10^{-4}\), and a source-RMS-relative threshold (Tables~\ref{tab:app-global-coordinate} and~\ref{tab:app-global-energy}).

\begin{table*}[t]
\centering
\small
\setlength{\tabcolsep}{4.2pt}
\resizebox{\textwidth}{!}{
\begin{tabular}{@{}lrrrrrrr@{}}
\toprule
& \multicolumn{2}{c}{Scale and support}
& \multicolumn{2}{c}{Update spectrum}
& \multicolumn{3}{c}{Source geometry} \\
\cmidrule(lr){2-3}\cmidrule(lr){4-5}\cmidrule(l){6-8}
Pair & Rel. norm & Visible spar. & Top-16 SVD & Stable rank & Prin. overlap & Low-mag. overlap & Spectral drift \\
 & (\%) & (\%) & energy (\%) & & (\%) & (\%) & \\
\midrule
\textcolor{distillgray}{Qwen-Math Distill} & \textcolor{distillgray}{11.9363} & \textcolor{distillgray}{3.06} & \textcolor{distillgray}{8.57} & \textcolor{distillgray}{82.75} & \textcolor{distillgray}{9.95} & \textcolor{distillgray}{10.09} & \textcolor{distillgray}{\(2.958{\times}10^{-2}\)} \\
\textcolor{distillgray}{R1-Qwen SFT (DeepScaleR)} & \textcolor{distillgray}{0.8280} & \textcolor{distillgray}{26.63} & \textcolor{distillgray}{15.34} & \textcolor{distillgray}{46.48} & \textcolor{distillgray}{8.76} & \textcolor{distillgray}{11.76} & \textcolor{distillgray}{\(1.782{\times}10^{-4}\)} \\
\textcolor{rlvrgray}{DeepScaleR RLVR} & \textcolor{rlvrgray}{0.0725} & \textcolor{rlvrgray}{77.89} & \textcolor{rlvrgray}{19.62} & \textcolor{rlvrgray}{16.07} & \textcolor{rlvrgray}{5.31} & \textcolor{rlvrgray}{29.73} & \textcolor{rlvrgray}{\(2.023{\times}10^{-5}\)} \\
R1-Qwen OPD & 0.0488 & 83.04 & 28.69 & 9.15 & 4.98 & 36.68 & \(1.387{\times}10^{-5}\) \\
\textcolor{distillgray}{R1-Qwen SFT (JustRL)} & \textcolor{distillgray}{0.7188} & \textcolor{distillgray}{29.02} & \textcolor{distillgray}{18.50} & \textcolor{distillgray}{27.60} & \textcolor{distillgray}{8.49} & \textcolor{distillgray}{12.23} & \textcolor{distillgray}{\(1.422{\times}10^{-4}\)} \\
\textcolor{rlvrgray}{JustRL RLVR} & \textcolor{rlvrgray}{0.4973} & \textcolor{rlvrgray}{34.68} & \textcolor{rlvrgray}{26.94} & \textcolor{rlvrgray}{12.35} & \textcolor{rlvrgray}{8.35} & \textcolor{rlvrgray}{14.44} & \textcolor{rlvrgray}{\(2.043{\times}10^{-3}\)} \\
R1-Qwen OPD (JustRL) & 0.1165 & 69.82 & 42.87 & 7.12 & 5.82 & 22.55 & \(3.612{\times}10^{-5}\) \\
MiniCPM OPD & 0.0689 & 81.65 & 26.81 & 12.32 & 5.41 & 36.95 & \(1.184{\times}10^{-5}\) \\
\textcolor{rlvrgray}{Qwen2.5-VL RLVR} & \textcolor{rlvrgray}{0.0387} & \textcolor{rlvrgray}{87.99} & \textcolor{rlvrgray}{20.46} & \textcolor{rlvrgray}{17.30} & \textcolor{rlvrgray}{4.70} & \textcolor{rlvrgray}{38.30} & \textcolor{rlvrgray}{\(9.929{\times}10^{-6}\)} \\
Qwen2.5-VL OPD & 0.1421 & 66.80 & 30.52 & 12.28 & 5.34 & 22.05 & \(3.194{\times}10^{-5}\) \\
Qwen3 OPD & 0.0445 & 79.03 & 19.69 & 20.30 & 4.46 & 32.06 & \(1.824{\times}10^{-5}\) \\
Qwen3 OPSD & 0.1732 & 63.74 & 29.49 & 8.52 & 5.70 & 20.66 & \(2.823{\times}10^{-5}\) \\
TCOD-B2F OPD & 0.0633 & 81.38 & 23.64 & 13.46 & 5.29 & 24.75 & \(1.202{\times}10^{-5}\) \\
\bottomrule
\end{tabular}
}
\caption{Checkpoint-delta statistics. Relative norm and visible sparsity aggregate over all matched tensors. The remaining metrics are unweighted medians over eligible 2D matrices. Principal and low-magnitude overlap use the visible support \(A_{10^{-5}}\) and 10\% source-coordinate masks. Spectral drift is normalized by the source spectrum. Gray text marks non-\textsc{OPD} references.}
\label{tab:global-delta}
\end{table*}

Although the JustRL \textsc{RLVR} checkpoint is atypically large and dense, using it as the teacher still produces a small, sparse, and off-principal \textsc{OPD} delta; a sparse teacher update is therefore not necessary for a sparse student update.

\subsection{Layer and module structure}

\begin{figure}[t]
\centering
\includegraphics[width=0.49\linewidth]{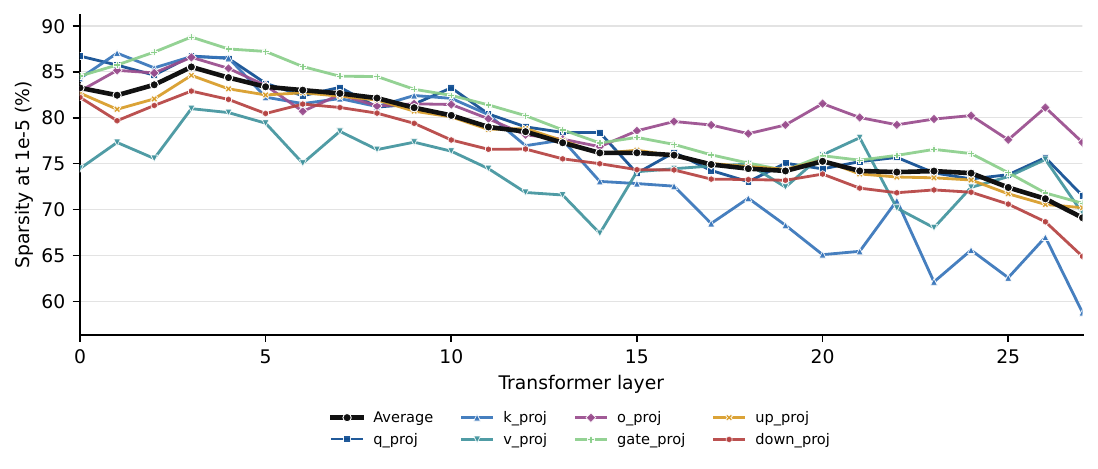}\hfill
\includegraphics[width=0.49\linewidth]{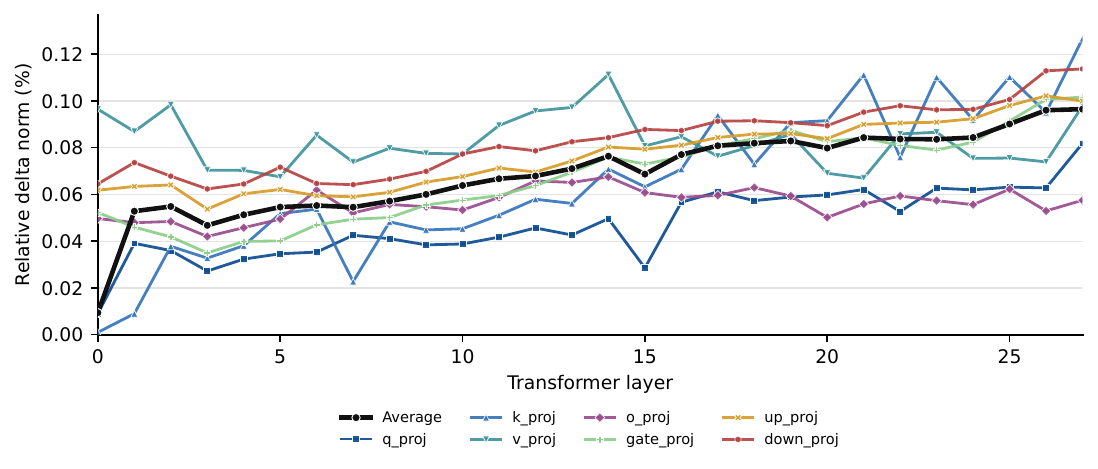}
\caption{Layerwise per-module update sparsity (left) and relative delta norm (right) for R1-Qwen OPD.}
\label{fig:layerwise-sparsity}
\end{figure}

Having established that \textsc{OPD} updates are globally small and sparse, we next ask how this structure is distributed across layers and modules.
For each layer--module group \(B\), Figure~\ref{fig:layerwise-sparsity} plots the same Frobenius-aggregated \(r(B)\) and coordinate-aggregated \(s_{10^{-5}}(B)\) defined above.

Figure~\ref{fig:layerwise-sparsity} shows that the small and sparse R1-Qwen \textsc{OPD} updates are distributed across layers and modules.
All major projection matrices receive threshold-visible updates throughout the network, yet their relative delta norms remain below \(0.14\%\), and most coordinates remain unchanged at the \(10^{-5}\) threshold.
No single module consistently dominates across layers.
This differs from the module-localized pattern reported for reasoning \textsc{SFT}, where \(o_{\mathrm{proj}}\) exhibits the largest or second-largest weight change and a distinct relative-change distribution~\citep{shao2026reasons}.

Component-wise statistics for the fully trained VLM are provided in Appendix~\ref{app:vlm-components}.

\subsection{Update-support overlap}

To test whether visible update supports are reused across changes in algorithm, teacher, or training data, we define thresholded supports \(A_j=\{i:|\Delta_i^{(j)}|>10^{-5}\}\) and compare them with the one-sided overlap scores
\begin{equation}
o_1=\frac{|A_1\cap A_2|}{|A_1|},
\qquad
o_2=\frac{|A_1\cap A_2|}{|A_2|}.
\end{equation}
These scores are directional because the support sizes can differ substantially; under independent supports, their respective baselines are \(|A_2|/N\) and \(|A_1|/N\), where \(N\) is the number of matched coordinates.

Across algorithm, teacher, and data variations, Table~\ref{tab:opd-rlvr-overlap} shows one-sided overlap at \(2.26\)--\(3.06\times\) its independent-support baseline.
For R1-Qwen, DeepScaleR \textsc{RLVR} and \textsc{OPD} attain \(51.81\%\) and \(67.54\%\) directional overlap, compared with \(16.96\%\) and \(22.11\%\) baselines.
The effect persists when varying the teacher within \textsc{OPD} and when holding the JustRL teacher fixed while changing the data from DAPO-Math-17K to OpenThoughts3.
Thus, the comparisons reveal \textbf{nonrandom support reuse} beyond what can be explained by the marginal support densities alone.

\begin{table*}[t]
\centering
\small
\setlength{\tabcolsep}{10pt}
\resizebox{\textwidth}{!}{
\begin{tabular}{@{}lllccc@{}}
\toprule
\shortstack[l]{Variation\\axis} & Model family & Setting & Random & Observed & \shortstack{Union\\sparsity} \\
\midrule
\multirow{2}{*}{Algorithm} & \multirow{2}{*}{R1-Qwen} & \(\mathcal{I}_1\): \textsc{RLVR} (DeepScaleR) & \(o_1\): 16.96 & \(o_1\): 51.81 (\(3.06\times\)) & \multirow{2}{*}{72.39} \\
 & & \(\mathcal{I}_2\): \textsc{OPD} (DeepScaleR teacher) & \(o_2\): 22.11 & \(o_2\): 67.54 (\(3.06\times\)) & \\
\midrule
\multirow{2}{*}{Algorithm} & \multirow{2}{*}{Qwen2.5-VL} & \(\mathcal{I}_1\): \textsc{RLVR} & \(o_1\): 33.20 & \(o_1\): 75.02 (\(2.26\times\)) & \multirow{2}{*}{63.80} \\
 & & \(\mathcal{I}_2\): \textsc{OPD} & \(o_2\): 12.01 & \(o_2\): 27.14 (\(2.26\times\)) & \\
\midrule
\multirow{2}{*}{Teacher} & \multirow{2}{*}{R1-Qwen} & \(\mathcal{I}_1\): \textsc{OPD} (DeepScaleR teacher) & \(o_1\): 30.18 & \(o_1\): 78.50 (\(2.60\times\)) & \multirow{2}{*}{66.18} \\
 & & \(\mathcal{I}_2\): \textsc{OPD} (JustRL teacher) & \(o_2\): 16.96 & \(o_2\): 44.11 (\(2.60\times\)) & \\
\midrule
\multirow{2}{*}{Data} & \multirow{2}{*}{R1-Qwen} & \(\mathcal{I}_1\): \textsc{OPD} (DAPO-Math-17K) & \(o_1\): 30.00 & \(o_1\): 71.90 (\(2.40\times\)) & \multirow{2}{*}{61.52} \\
 & & \(\mathcal{I}_2\): \textsc{OPD} (OpenThoughts3) & \(o_2\): 30.18 & \(o_2\): 72.33 (\(2.40\times\)) & \\
\bottomrule
\end{tabular}
}
\caption{Overlap between visible update supports at \(|\Delta_i|>10^{-5}\). Within each paired comparison, \(o_1\) and \(o_2\) are one-sided overlap scores, and Random gives the corresponding independent-support baseline. Union sparsity is the fraction of coordinates updated by neither run and is shared by the two rows in each pair.}
\label{tab:opd-rlvr-overlap}
\end{table*}

\subsection{Fine-tuning the subnetwork alone suffices in \textsc{OPD}}

We test whether training on the discovered \textsc{OPD} support is sufficient to recover full-training performance.
Given a source checkpoint \(\theta_0\) and a full \textsc{OPD} checkpoint \(\theta_{\mathrm{full}}\), we define the full-\textsc{OPD} mask
\begin{equation}
    m_i = \mathbb{1}\{|\theta_{\mathrm{full},i} - \theta_{0,i}| > \tau\},
\end{equation}
where \(\tau=10^{-5}\) after subtracting the stored checkpoint weights.
We restart from \(\theta_0\) with the same training configuration and restrict updates to either the learned \textsc{OPD} support, an \textsc{RLVR} support, or a density-matched random support; full \textsc{OPD} provides the reference.

\begin{figure}[t]
\centering
\includegraphics[width=0.49\linewidth]{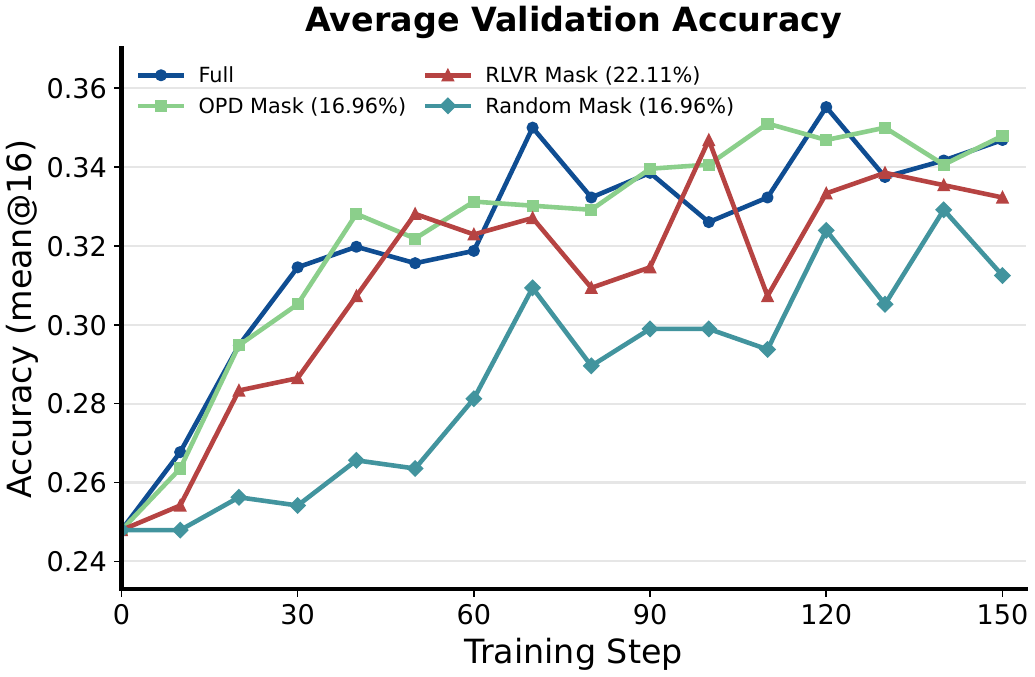}\hfill
\includegraphics[width=0.49\linewidth]{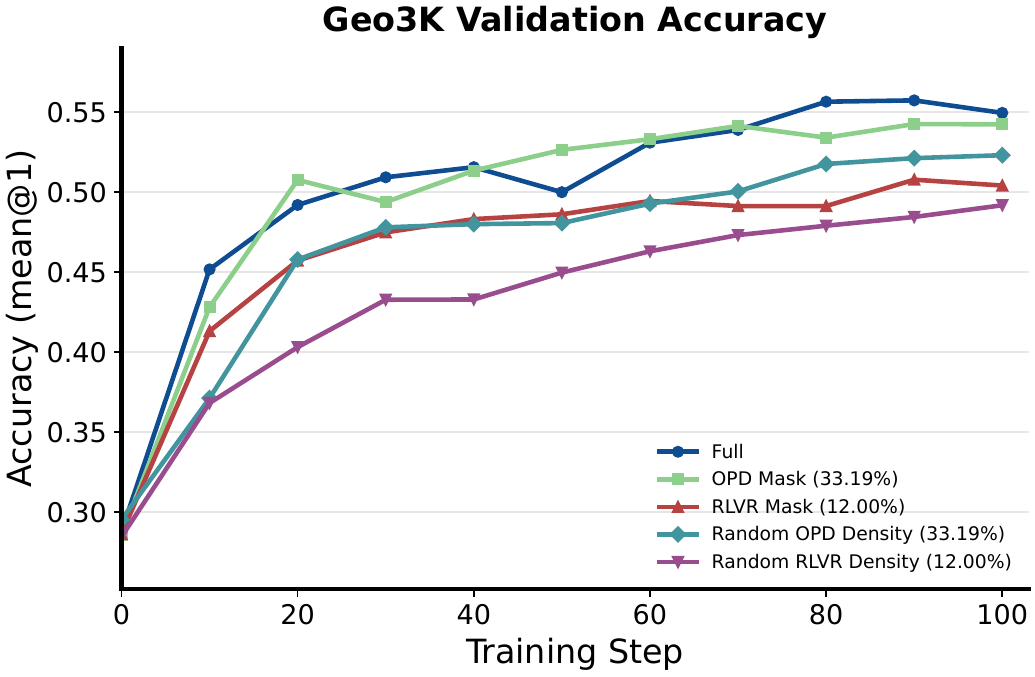}
\caption{Subnetwork-masked \textsc{OPD} across LLM and VLM settings. Percentages in parentheses denote the fraction of parameters permitted to update under each mask. Left: average validation accuracy over AIME24 and AIME25 for R1-Qwen. Right: Geo3K validation accuracy for Qwen2.5-VL. The learned \textsc{OPD} masks closely track full training, while density-matched random masks are weaker. Benchmark-wise AIME curves are provided in Appendix~\ref{app:benchmark-wise-functional}.}
\label{fig:functional-average}
\end{figure}

Across the R1-Qwen reasoning and Qwen2.5-VL Geo3K settings, training on the learned \textsc{OPD} support \textbf{closely tracks full training} and consistently outperforms its density-matched random baseline (Figure~\ref{fig:functional-average}).
For R1-Qwen, the learned-mask run peaks at \(35.10\%\) mean@16 accuracy, close to full training at \(35.52\%\) and above the random-mask peak of \(32.92\%\).
The Qwen2.5-VL experiment shows the same ordering: the learned-mask, full-training, and density-matched random runs peak at \(54.24\%\), \(55.72\%\), and \(52.30\%\), respectively.
The \textsc{RLVR} supports also permit successful training but are generally weaker than the corresponding \textsc{OPD} supports, consistent with their substantial but incomplete overlap in Table~\ref{tab:opd-rlvr-overlap}.
The selected coordinates therefore matter beyond imposing a sparse parameter budget.

Operationally, the \textsc{OPD} delta defines a sparse task vector whose support is sufficient to nearly recover the measured performance gains \citep{ilharco2023editing}.
Unlike a classical lottery ticket uncovered by pruning a dense network \citep{frankle2019lottery}, this support emerges directly from the checkpoint delta.

\section{Geometry of \textsc{OPD} updates}
\label{sec:geometry}

\subsection{Spectral concentration despite full numerical rank}

For each 2D update matrix with exact singular values \(\sigma_1\geq\cdots\geq\sigma_m\), we measure
\begin{equation}
\begin{aligned}
e_k
&=\frac{\sum_{i=1}^{k}\sigma_i^2}{\sum_{i=1}^{m}\sigma_i^2},
\\
\operatorname{srank}(\Delta W)
&=\frac{\|\Delta W\|_F^2}{\sigma_1^2},
\\
\overline{\operatorname{rank}}_\tau(\Delta W)
&=\frac{|\{i:\sigma_i>\tau\sigma_1\}|}{m},
\qquad \tau=10^{-5}.
\end{aligned}
\end{equation}
Higher \(e_k\) and lower stable rank indicate stronger energy concentration, whereas \(\overline{\operatorname{rank}}_\tau=1\) means numerical full rank at tolerance \(\tau\).
Reported spectral values are medians over matrices for which exact SVD is available.

\textsc{OPD} update energy is \textbf{concentrated in a small leading singular subspace}.
Across all seven pairs, top-16 energy is higher and stable rank is lower than in the fixed-data \textsc{SFT} and offline distillation references (Figure~\ref{fig:static-summary} and Table~\ref{tab:global-delta}).
In the matched R1-Qwen/DeepScaleR comparison, top-16 energy increases from \(15.34\%\) under \textsc{SFT} to \(28.69\%\) under \textsc{OPD}, while stable rank decreases from \(46.48\) to \(9.15\).
The similar profile of DeepScaleR \textsc{RLVR} further indicates that this concentration is not specific to dense teacher supervision.
At the relative tolerance \(\tau=10^{-5}\), the median normalized numerical rank is \(100\%\) for every \textsc{OPD} pair (Table~\ref{tab:app-spectral}), so the concentration reflects dominant leading directions rather than rank truncation.
The intermediate-checkpoint analysis in Appendix~\ref{app:training-trajectory} complements this picture: in the two JustRL-teacher runs, \textsc{SFT} updates spread across more coordinates and singular directions as training proceeds, whereas \textsc{OPD} support expands while median stable rank remains between \(6.17\) and \(6.80\) over the observed checkpoints.

\subsection{Functional dominance of the top-rank update subspace}
\label{subsec:functional-top-rank}

The concentrated spectra above suggest that the leading singular directions may carry most of the behavioral effect of an update.
We test this hypothesis through a functional intervention on the matched R1-Qwen runs trained with the DeepScaleR teacher.
For every self-attention and MLP weight matrix \(p\), we compute the exact SVD of \(\Delta W_p\), retain its leading
\(q_p=\lceil 0.01\min(d_{\mathrm{out}},d_{\mathrm{in}})\rceil\) components to obtain \(\Delta W_{p,1\%}\), and restore the original per-matrix update norm:
\begin{equation}
    \widehat{\Delta W}_{p,1\%}
    =
    \frac{\|\Delta W_p\|_F}
         {\|\Delta W_{p,1\%}\|_F}
    \Delta W_{p,1\%}.
\end{equation}
The intervened model adds these reconstructed updates to the source checkpoint and keeps all other parameters at their source values.
We quantify recovery relative to the common source model as
\begin{equation}
    R_{\mathrm{gain}}
    =
    \frac{A_{\mathrm{Top}\text{-}1\%}-A_{\mathrm{Base}}}
         {A_{\mathrm{Full}}-A_{\mathrm{Base}}},
\end{equation}
where \(A\) is the mean score over 16 sampled responses per problem.
Further implementation and update-energy diagnostics are provided in Appendices~\ref{app:additional-static} and~\ref{app:top1-svd-reconstruction}.

\begin{table}[t]
\centering
\small
\begin{tabular*}{\linewidth}{@{\extracolsep{\fill}}llrrrr@{}}
\toprule
Regime & Benchmark & Base & Full & Top-1\% & \(R_{\mathrm{gain}}\) (\%) \\
\midrule
\textsc{OPD} & AIME24 & 29.17 & 39.38 & \textbf{41.46} & 120.41 \\
\textsc{OPD} & AIME25 & 24.38 & 27.08 & \textbf{27.50} & 115.38 \\
\midrule
\textsc{SFT} & AIME24 & 29.17 & 38.33 & 1.67 & -300.00 \\
\textsc{SFT} & AIME25 & 24.38 & 28.96 & 1.46 & -500.00 \\
\bottomrule
\end{tabular*}
\caption{Functional recovery from the norm-matched Top-1\% update subspace on the matched DeepScaleR experiments. All scores are AIME mean scores (\%). \(R_{\mathrm{gain}}\) measures recovery of the improvement over the common Base; negative values indicate performance below Base.}
\label{tab:top1-functional-recovery}
\end{table}

Table~\ref{tab:top1-functional-recovery} reveals a sharp functional separation under the same source model, teacher, and reconstruction rule.
For \textsc{OPD}, the Top-1\% reconstruction retains the measured reasoning gains, with gain-recovery ratios of \(120.41\%\) and \(115.38\%\) on AIME24 and AIME25, respectively.
The corresponding \textsc{SFT} reconstruction instead falls below \(2\%\) on both benchmarks and below the Base model.
A norm-matched reconstruction in the leading update subspace is therefore sufficient to reproduce the measured \textsc{OPD} reasoning gains, whereas the same intervention substantially degrades \textsc{SFT} performance.

\subsection{Off-principal movement from source weights}

We characterize source-relative geometry using source-spectrum drift and overlap between the visible update support and source-derived coordinate masks.
For each source matrix \(W_{\mathrm{src}}=U\Sigma V^\top\), let \(m=\min(d_{\mathrm{in}},d_{\mathrm{out}})\) and \(R_k=U_k\Sigma_kV_k^\top\) be its rank-\(k\) reconstruction.
The principal mask \(M_{\mathrm{prin}}^\rho\) contains the \(\lceil\rho|W|\rceil\) coordinates with largest \(|R_{k,i}|\), whereas the low-magnitude mask \(M_{\mathrm{low}}^\rho\) contains the same number of coordinates with smallest \(|W_{\mathrm{src},i}|\).
For the visible update support \(A_{10^{-5}}=\{i:|\Delta W_i|>10^{-5}\}\), source-mask overlap is
\begin{equation}
    o(M)=\frac{|A_{10^{-5}}\cap M|}{|A_{10^{-5}}|}.
\end{equation}
We additionally measure the relative drift of the source singular-value spectrum:
\begin{equation}
    d_{\mathrm{spec}}(W)=
    \frac{\|\sigma(W_{\mathrm{trained}})-\sigma(W_{\mathrm{src}})\|_2}
         {\|\sigma(W_{\mathrm{src}})\|_2}.
\end{equation}
The main results set \(\rho=0.1\), \(k=\max(1,\lfloor\rho m\rfloor)\), and report unweighted medians over eligible 2D matrices.

At checkpoint precision, \textsc{OPD} visible supports consistently show \textbf{reduced overlap with source-principal coordinate masks and enrichment in low-magnitude source coordinates} (Table~\ref{tab:global-delta}).
For R1-Qwen \textsc{OPD}, principal-mask overlap is \(4.98\%\), below the \(10\%\) independent-support baseline, while low-magnitude-mask overlap reaches \(36.68\%\).
The fixed-data \textsc{SFT} and offline distillation references remain much closer to independent-support overlap, whereas DeepScaleR \textsc{RLVR} exhibits the same qualitative bias as \textsc{OPD}.
\textsc{OPD} also produces substantially less source-spectrum drift than the fixed-data references.
For the same R1-Qwen comparison, median drift is \(1.387\times10^{-5}\) under \textsc{OPD} versus \(1.782\times10^{-4}\) under \textsc{SFT}.
Together, these results indicate localized writing away from source-principal coordinates without broadly changing the source singular-value spectrum.

\section{Discussion and future work}

Taken together, our results show that dense teacher feedback can yield sparse, concentrated checkpoint updates with functional value, extending weight-space observations from \textsc{RLVR} to \textsc{OPD}.
The learned-support retraining results motivate support selection for sparse fine-tuning, while the reconstruction results motivate studying singular-subspace initialization for \textsc{LoRA} variants such as MiLoRA and PiSSA \citep{wang2024milora,meng2024pissa}.
The contrasting response of \textsc{OPD} and \textsc{SFT} to the same Top-1\% reconstruction makes the training regime relevant to the choice of adaptation subspace.
On the optimizer side, comparing AdamW with Muon and its high-pass variants could test how spectral normalization changes \textsc{OPD}'s update concentration and functional recovery \citep{jordan2024muon,fan2026rethinking}.
Jointly tracking these quantities during training would connect optimizer design to the sparse supports and leading directions identified by our interventions.

Our analyses span language and vision-language models, math reasoning, and agent training, with functional interventions on R1-Qwen and Qwen2.5-VL math reasoning.
Future work should test broader scales and domains, trace training dynamics, connect weight-space geometry to behavioral change and forgetting \citep{shenfeld2025rl}, and develop geometry-aware adaptation and optimization.

\section{Conclusion}

We show that \textsc{OPD} produces small, coordinate-sparse, spectrally concentrated, off-principal, and spectrum-preserving updates whose restricted subnetworks and leading singular components retain most reasoning gains.
These findings establish the functional relevance of \textsc{OPD} update structure and motivate parameter-efficient adaptation through coordinate selection and low-dimensional update subspaces.

\clearpage
\bibliographystyle{colm2026_conference}
\bibliography{colm2026_conference}

\newpage
\appendix

\section{Related work}
\label{app:related-work}

\paragraph{On-policy distillation.}
Early \textsc{OPD} formulations emphasize the exposure-mismatch problem in offline sequence-level distillation and train the student on its own sampled trajectories with dense teacher feedback \citep{agarwal2024onpolicy,gu2024minillm}.
Recently, this idea has become a practical component of large language model post-training \citep{song2026surveyopd}, including \textit{large-to-small distillation} \citep{yang2025qwen3,lu2025onpolicydistillation,li2026rethinking}, \textit{capability consolidation with multiple experts} \citep{xiaomi2026mimov2flash,glm5team2026glm5,yang2026nemotron,deepseekai2026deepseekv4,kimiteam2026kimik3openfrontier,openbmb2026minicpm5}, and \textit{on-policy self-distillation from privileged information} \citep{zhao2026selfdistilledreasoner,hubotter2026rlselfdistillation,ye2026policy}.
Our work complements these training methods by characterizing \textsc{OPD} parameter updates and testing the functional role of their structure.

\paragraph{Parameter geometry of post-training.}
A second line of related work analyzes the parameter geometry of post-training.
\citet{mukherjee2025reinforcementlearningfinetunessmall} show that \textsc{RLVR} fine-tunes small subnetworks in large language models, while \citet{zhu2025pathtakenrlvrprovably} argue that \textsc{RLVR} learns away from the principal directions of source weights.
\citet{cai2025predictabilityreinforcementlearningdynamics} further study reinforcement-learning-induced parameter dynamics and find that dominant low-rank directions can predict later reasoning improvements.
We ask whether these properties persist when the sparse reward signal is replaced by dense teacher supervision, which places \textsc{OPD} between conventional dense-supervised fine-tuning and sparse-reward on-policy training.

\paragraph{Adaptation and optimization methods.}
The spectral and source-relative coordinate results also relate to parameter-efficient adaptation and optimizer design.
Low-rank adaptation methods such as \textsc{LoRA} exploit the concentration of fine-tuning updates in a low-dimensional additive subspace \citep{hu2022lora,schulman2025lora}.
MiLoRA and PiSSA initialize adapters from minor and principal singular components of the source weights, respectively \citep{wang2024milora,meng2024pissa}.
Our results motivate testing how such source-based initialization choices interact with the concentrated update directions observed in \textsc{OPD}.
Orthogonal fine-tuning methods such as \textsc{OFT}/\textsc{BOFT} instead parameterize adaptation through orthogonal transformations, aiming to preserve hyperspherical energy and pretrained representation structure while changing the model's function \citep{liu2024boft}.
Muon is relevant from the optimizer side because it treats hidden-layer weights as matrices and orthogonalizes the momentum update before applying it \citep{jordan2024muon,liu2025muon}.
Recent work argues that this spectral normalization can fail in \textsc{RLVR} post-training and proposes high-pass remedies, making these variants a more plausible starting point for \textsc{OPD} than vanilla Muon \citep{fan2026rethinking}.
These methods motivate future \textsc{OPD}-native optimization strategies.

\section{Post-training objectives for LLMs}
\label{app:post-training-objectives}

We summarize the training objectives that define the regimes compared in this paper.
Let \(x\) denote a prompt, \(y=(y_1,\ldots,y_T)\) a response, \(\pi_\theta\) the trainable policy, \(\pi_T\) a teacher policy, and \(\pi_{\mathrm{ref}}\) a reference policy used for KL regularization when applicable.

\paragraph{SFT and sequence-level KD.}
Supervised fine-tuning (\textsc{SFT}) optimizes the negative log-likelihood of fixed target responses:
\begin{equation}
    \mathcal{L}_{\mathrm{SFT}}(\theta)
    =
    -\mathbb{E}_{(x,y^\star)\sim\mathcal{D}}
    \sum_{t=1}^{T}
    \log \pi_\theta(y_t^\star \mid x,y_{<t}^\star).
\end{equation}
Sequence-level knowledge distillation (\textsc{SeqKD}) uses the same maximum-likelihood objective, but the target response \(y^T\) is generated by a teacher model rather than drawn from human annotations \citep{hinton2015distilling,kim2016sequence}:
\begin{equation}
    \mathcal{L}_{\mathrm{SeqKD}}(\theta)
    =
    -\mathbb{E}_{x\sim\mathcal{D},\, y^T\sim\pi_T(\cdot|x)}
    \sum_{t=1}^{T}
    \log \pi_\theta(y_t^T \mid x,y_{<t}^T).
\end{equation}
Both objectives are dense in the sense that every target token supplies a supervised learning signal, but the trajectories are offline-collected and fixed before the student update.

\paragraph{RLVR and GRPO.}
Reinforcement learning with verifiable rewards (\textsc{RLVR}) samples responses from the current or recent policy and optimizes a scalar reward \(r(x,y)\), often with a KL penalty to keep the policy close to \(\pi_{\mathrm{ref}}\):
\begin{equation}
    \mathcal{J}_{\mathrm{RLVR}}(\theta)
    =
    \mathbb{E}_{\substack{x\sim\mathcal{D}\\y\sim\pi_\theta(\cdot|x)}}
    \left[
    r(x,y)
    -
    \beta
    \sum_{t=1}^{T}
    D_{\mathrm{KL}}\!\left(
        \pi_\theta(\cdot|x,y_{<t})
        \,\|\,
        \pi_{\mathrm{ref}}(\cdot|x,y_{<t})
    \right)
    \right].
\end{equation}
GRPO is a seminal \textsc{RLVR} method that removes the learned critic by sampling a group
of responses
\(\{y_i\}_{i=1}^{G}\) for the same prompt and normalizing their rewards within the group
\citep{shao2024deepseekmath}.
With
\(\hat A_i=(r_i-\mathrm{mean}(\{r_j\}_{j=1}^{G}))/(\mathrm{std}(\{r_j\}_{j=1}^{G})+\epsilon)\)
and
\(\rho_{i,t}(\theta)=\pi_\theta(y_{i,t}|x,y_{i,<t})/\pi_{\mathrm{old}}(y_{i,t}|x,y_{i,<t})\),
a clipped GRPO objective can be written as
\begin{equation}
\begin{aligned}
    \mathcal{J}_{\mathrm{GRPO}}(\theta)
    &=
    \mathbb{E}_{x,\{y_i\}_{i=1}^{G}}
    \left[
    \frac{1}{G}
    \sum_{i=1}^{G}
    \frac{1}{|y_i|}
    \sum_{t=1}^{|y_i|}
    \left[g_{i,t}(\theta)-\beta k_{i,t}(\theta)\right]
    \right],
    \\
    g_{i,t}(\theta)
    &=
    \min\!\Bigl(
        \rho_{i,t}(\theta)\hat A_i,
        \mathrm{clip}\!\left(
            \rho_{i,t}(\theta),
            1-\epsilon,
            1+\epsilon
        \right)\hat A_i
    \Bigr),
    \\
    k_{i,t}(\theta)
    &=
    D_{\mathrm{KL}}\!\left(
        \pi_\theta(\cdot|x,y_{i,<t})
        \,\|\,
        \pi_{\mathrm{ref}}(\cdot|x,y_{i,<t})
    \right).
\end{aligned}
\end{equation}
RLVR therefore uses on-policy or near-on-policy samples but usually receives sparse sequence-level
reward feedback.

\paragraph{OPD.}
\textsc{OPD} combines on-policy student rollouts with dense teacher feedback.
One style, which we call GKD-style \textsc{OPD}, applies a token-level teacher-student $f$-divergence on prefixes generated by the student \citep{agarwal2024onpolicy}:
\begin{equation}
    \mathcal{L}_{\mathrm{OPD}\text{-}\mathrm{GKD}}(\theta)
    =
    \mathbb{E}_{\substack{x\sim\mathcal{D}\\y\sim\pi_\theta(\cdot|x)}}
    \sum_{t=1}^{|y|}
    D_f\!\left(
        \pi_T(\cdot|x,y_{<t})
        \,\|\, 
        \pi_\theta(\cdot|x,y_{<t})
    \right),
\end{equation}
where \(D_f\) is typically an $f$-divergence over token distributions.
Another style, which we call PG-style \textsc{OPD}, treats distillation as policy optimization on student-generated samples \citep{gu2024minillm}.
An abstract form is the reverse-KL sequence objective
\begin{equation}
    \mathcal{L}_{\mathrm{OPD}\text{-}\mathrm{PG}}(\theta)
    =
    \mathbb{E}_{\substack{x\sim\mathcal{D}\\y\sim\pi_\theta(\cdot|x)}}
    \sum_{t=1}^{|y|}
    \left[
        \log \pi_\theta(y_t|x,y_{<t})
        -
        \log \pi_T(y_t|x,y_{<t})
    \right],
\end{equation}
which can be optimized with policy-gradient estimators because the sampled trajectory depends on the student policy.
This PG view is connected to reverse-KL distillation through a simple score-function identity.
For a fixed prefix \(c=(x,y_{<t})\), define the sampled-token estimator
\(k_1(a;c)=\log \pi_\theta(a|c)-\log \pi_T(a|c)\).
The full-vocabulary reverse KL at this prefix is
\begin{equation}
    D_{\mathrm{KL}}\!\left(
        \pi_\theta(\cdot|c)
        \,\|\,
        \pi_T(\cdot|c)
    \right)
    =
    \sum_a
    \pi_\theta(a|c)
    k_1(a;c).
\end{equation}
Its gradient is
\begin{equation}
\begin{aligned}
    \nabla_\theta
    D_{\mathrm{KL}}\!\left(
        \pi_\theta(\cdot|c)
        \,\|\,
        \pi_T(\cdot|c)
    \right)
    &=
    \nabla_\theta
    \sum_a
    \pi_\theta(a|c)
    k_1(a;c)
    \\
    &=
    \sum_a
    \nabla_\theta \pi_\theta(a|c)
    k_1(a;c)
    +
    \sum_a
    \pi_\theta(a|c)
    \nabla_\theta k_1(a;c)
    \\
    &=
    \mathbb{E}_{a\sim \pi_\theta(\cdot|c)}
    \left[
        k_1(a;c)
        \nabla_\theta\log\pi_\theta(a|c)
    \right]
    +
    \sum_a
    \nabla_\theta\pi_\theta(a|c)
    \\
    &=
    \mathbb{E}_{a\sim \pi_\theta(\cdot|c)}
    \left[
        k_1(a;c)
        \nabla_\theta\log\pi_\theta(a|c)
    \right].
\end{aligned}
\end{equation}
Here we use
\(\nabla_\theta\pi_\theta(a|c)=\pi_\theta(a|c)\nabla_\theta\log\pi_\theta(a|c)\)
and \(\nabla_\theta k_1(a;c)=\nabla_\theta\log\pi_\theta(a|c)\), since the
prefix \(c\) and the teacher \(\pi_T\) are fixed.
The final equality uses \(\sum_a \pi_\theta(a|c)=1\), so
\(\sum_a \nabla_\theta\pi_\theta(a|c)=\nabla_\theta 1=0\).
Thus, if \(k_1\) is detached and \(-k_1(a;c)\) is used as the policy-gradient
advantage, the score-function loss
\(\mathcal{L}_{\mathrm{PG}}(a;c)=\operatorname{sg}[k_1(a;c)]\log \pi_\theta(a|c)\)
has gradient
\(k_1(a;c)\nabla_\theta\log \pi_\theta(a|c)\), giving an unbiased single-sample
estimator of the reverse-KL gradient up to practical modifications such as clipping,
off-policy importance ratios, and advantage normalization.

Both \textsc{OPD} styles differ from \textsc{SeqKD} by training on student-generated trajectories, and differ from \textsc{RLVR} by replacing sparse scalar rewards with dense teacher-derived feedback.

\section{Additional details}

\subsection{Model pairs and training details}
\label{app:model-pairs-training}

\begin{table*}[t]
\centering
\footnotesize
\begin{tabularx}{\textwidth}{@{}YYYYY@{}}
\toprule
Pair & Source model & Trained model & Training regime / learning signal & Data \\
\midrule
\textcolor{distillgray}{Qwen-Math Distill} & \textcolor{distillgray}{Qwen2.5-Math-1.5B} & \textcolor{distillgray}{R1-Qwen} & \textcolor{distillgray}{Sequence distillation; DeepSeek-R1 teacher} & \textcolor{distillgray}{Public checkpoint} \\
\textcolor{distillgray}{R1-Qwen SFT (DeepScaleR)} & \textcolor{distillgray}{R1-Qwen} & \textcolor{distillgray}{R1-Qwen SFT} & \textcolor{distillgray}{Full-parameter \textsc{SFT} on offline DeepScaleR responses} & \textcolor{distillgray}{DAPO-Math-17K} \\
\textcolor{rlvrgray}{DeepScaleR RLVR} & \textcolor{rlvrgray}{R1-Qwen} & \textcolor{rlvrgray}{DeepScaleR-1.5B} & \textcolor{rlvrgray}{\textsc{GRPO}; verifiable rewards} & \textcolor{rlvrgray}{Public RLVR checkpoint} \\
R1-Qwen OPD & R1-Qwen & R1-Qwen OPD & \textsc{OPD}; DeepScaleR-1.5B teacher & DAPO-Math-17K \\
\textcolor{distillgray}{R1-Qwen SFT (JustRL)} & \textcolor{distillgray}{R1-Qwen} & \textcolor{distillgray}{R1-Qwen SFT} & \textcolor{distillgray}{Full-parameter \textsc{SFT} on offline JustRL responses} & \textcolor{distillgray}{DAPO-Math-17K} \\
\textcolor{rlvrgray}{JustRL RLVR} & \textcolor{rlvrgray}{R1-Qwen} & \textcolor{rlvrgray}{JustRL} & \textcolor{rlvrgray}{\textsc{GRPO}; simple recipe} & \textcolor{rlvrgray}{Public RLVR checkpoint} \\
R1-Qwen OPD (JustRL) & R1-Qwen & R1-Qwen OPD & \textsc{OPD}; JustRL teacher & DAPO-Math-17K \\
R1-Qwen OPD (OpenThoughts3) & R1-Qwen & R1-Qwen OPD & \textsc{OPD}; JustRL teacher & Subsets of OpenThoughts3 \\
MiniCPM OPD & MiniCPM5-1B-SFT & MiniCPM5-1B & \textsc{OPD}; MiniCPM5 RL+OPD recipe & Public OPD checkpoint \\
\textcolor{rlvrgray}{Qwen2.5-VL RLVR} & \textcolor{rlvrgray}{Qwen2.5-VL-3B-Instruct} & \textcolor{rlvrgray}{Qwen2.5-VL RLVR} & \textcolor{rlvrgray}{\textsc{GRPO}; verifiable rewards} & \textcolor{rlvrgray}{Geo3K} \\
Qwen2.5-VL OPD & Qwen2.5-VL-3B-Instruct & Qwen2.5-VL OPD & \textsc{OPD}; NoisyRollout-7B teacher & Geo3K \\
Qwen3 OPD & Qwen3-1.7B-Base & Qwen3-1.7B-Base OPD & \textsc{OPD}; Qwen3-4B-Base-GRPO teacher & DAPO-Math-17K \\
Qwen3 OPSD & Qwen3-4B & Qwen3 OPSD & \textsc{OPSD}; self-distillation with privileged traces & OpenThoughts-Math-30K subsets \\
TCOD-B2F OPD & Qwen2.5-7B-Instruct & TCOD-B2F & \textsc{OPD}; temporal curriculum & ALFWorld \\
\bottomrule
\end{tabularx}
\caption{The 13 model pairs in the checkpoint-delta summary and the additional OpenThoughts3 pair for support-overlap and training-trajectory analyses. The comparisons span offline distillation, \textsc{SFT}, \textsc{RLVR}, \textsc{OPD}, and \textsc{OPSD} settings across LLM and VLM checkpoints.}
\label{tab:model-pairs}
\end{table*}

Table~\ref{tab:model-pairs} lists the checkpoint pairs used in the delta analysis.
These pairs cover LLMs with PG-style \textsc{OPD} updates and VLMs with GKD-style \textsc{OPD} updates, and they span the three use cases of \textsc{OPD} discussed in Section~\ref{app:related-work}.

For training and evaluation, we use the public HybridFlow implementation in verl v0.8.0 \citep{sheng2024hybridflow} and the public \textsc{OPD} and \textsc{OPSD} implementations where applicable.\footnote{\url{https://github.com/verl-project/verl}, \url{https://github.com/thunlp/OPD}, and \url{https://github.com/siyan-zhao/OPSD}.}
The R1-Qwen \textsc{OPD} runs use DAPO-Math-17K \citep{yu2025dapo} or OpenThoughts3 \citep{guha2025openthoughts}, the VLM \textsc{GRPO} and \textsc{OPD} runs use Geo3K \citep{lu2021intergps}, the Qwen3-1.7B \textsc{OPD} pair is a public checkpoint distilled from Qwen3-4B-Base-GRPO, and the \textsc{OPSD} run uses OpenThoughts-Math-30K subsets.
For the TCOD pair, we use Qwen2.5-7B-Instruct and the public \texttt{kolerk/tcod\_7b\_b2f} checkpoint \citep{wang2026tcod}, covering 7.616 billion parameters in 339 matched tensors and 198 two-dimensional matrices.
Tables~\ref{tab:R1-Qwen-opd-config} and~\ref{tab:qwen25vl-opd-config} summarize the core training configurations for the controlled R1-Qwen and Qwen2.5-VL \textsc{OPD} runs.

\paragraph{Configuration selection and reproducibility.}
Each experiment uses one fixed configuration based on its corresponding implementation, as reported in the configuration tables.
For verl-based training, we use the default seed configuration of verl v0.8.0.
Our training runs use NVIDIA A800 GPUs.
All AIME24 and AIME25 evaluations use temperature \(0.7\), top-\(p=0.95\), and 16 responses per problem.

\paragraph{Offline SFT references.}
To contrast on-policy distillation with fixed offline teacher supervision, we construct two \textsc{SFT} references from the same R1-Qwen source checkpoint.
We use DeepScaleR-1.5B-Preview and JustRL-DeepSeek-1.5B separately to generate one response per DAPO-Math-17K prompt.
The rollout pipeline uses temperature \(1.0\), top-\(p=0.95\), top-\(k=-1\), and a maximum response length of 7,168 tokens.
Responses without a boxed answer or with degenerate repetition are rejected, with at most three generation attempts per prompt; the accepted prompt--response pairs then form fixed supervised corpora \citep{li2026rethinking}.
We full-parameter fine-tune R1-Qwen with LLaMA-Factory v0.9.5 \citep{zheng2024llamafactory} using the Qwen template and the settings in Table~\ref{tab:R1-Qwen-sft-config}.

\begin{table*}[t]
\centering
\small
\begin{tabularx}{\textwidth}{@{}lX@{}}
\toprule
Setting & Value \\
\midrule
Student model & DeepSeek-R1-Distill-Qwen-1.5B (R1-Qwen) \\
Offline teachers & DeepScaleR-1.5B-Preview; JustRL-DeepSeek-1.5B \\
Training data & One accepted teacher response per DAPO-Math-17K prompt \\
Fine-tuning & Full-parameter \textsc{SFT}; Qwen template; sequence-length cutoff 8,192 \\
Batching / precision & Per-device batch size 4 on four GPUs; gradient accumulation 1; BF16 \\
Optimization & 3 epochs; Learning rate \(10^{-5}\); cosine schedule; warmup ratio 0.05 \\
\bottomrule
\end{tabularx}
\caption{Training configuration for the two fixed-data R1-Qwen \textsc{SFT} references.}
\label{tab:R1-Qwen-sft-config}
\end{table*}

\begin{table*}[t]
\centering
\small
\begin{tabularx}{\textwidth}{@{}lX@{}}
\toprule
Setting & Value \\
\midrule
Student model & DeepSeek-R1-Distill-Qwen-1.5B \\
Teacher model & DeepScaleR-1.5B or JustRL-DeepSeek-1.5B \\
Training / validation data & DAPO-Math-17K or OpenThoughts3 / AIME24 and AIME25 \\
Distillation loss & \(k_1\) policy-gradient style \textsc{OPD}; no task rewards \\
Rollout backend & vLLM rollout, \(n=4\) responses per prompt \\
Batching & train batch size 128; PPO mini-batch size 32 \\
Sequence lengths & max prompt 1024; max train response 7168; max validation response 15360 \\
Optimizer & AdamW, learning rate \(10^{-6}\) \\
Additional regularization & No additional reference-model KL loss \\
Distributed setup & 1 node, 4 actor GPUs per node, 2 teacher GPUs per node \\
Training schedule & 1 epoch; evaluate every 10 steps \\
\bottomrule
\end{tabularx}
\caption{Core training configuration for the R1-Qwen \textsc{OPD} runs. The OpenThoughts3 run changes the training data while retaining the JustRL teacher.}
\label{tab:R1-Qwen-opd-config}
\end{table*}

\begin{table*}[t]
\centering
\small
\begin{tabularx}{\textwidth}{@{}lX@{}}
\toprule
Setting & Value \\
\midrule
Student model & Qwen2.5-VL-3B-Instruct \\
Teacher model & NoisyRollout-Geo3K-7B \\
Training / validation data & Geo3K train / test split \\
Distillation loss & Forward-KL top-\(k\) \textsc{OPD} with \(k=32\); no task rewards \\
Rollout backend & vLLM rollout, \(n=1\) response per prompt \\
Batching & train batch size 128; PPO mini-batch size 32\\
Sequence lengths & max prompt 1024; max response 2048 \\
Optimizer & AdamW, learning rate \(10^{-6}\) \\
Additional regularization & No additional reference-model KL loss \\
Distributed setup & 1 node, 4 actor GPUs per node, 2 teacher GPUs per node \\
Training schedule & 10 epochs; evaluate every 10 steps \\
\bottomrule
\end{tabularx}
\caption{Core training configuration for Qwen2.5-VL \textsc{OPD}.}
\label{tab:qwen25vl-opd-config}
\end{table*}

\paragraph{Random-mask baselines.}
For both R1-Qwen and Qwen2.5-VL, the density-matched random baseline matches only a reference mask's global active-parameter fraction.
In both settings, active coordinates are allocated approximately in proportion to each floating-point tensor's size and sampled uniformly within fixed-size chunks.
We generate one random mask per setting using seed 42.

\subsection{Delta metrics and their interpretation}
\label{app:delta-metrics}

The checkpoint analysis in Section~\ref{sec:static-setup} uses the following metric definitions and aggregation rules.
\paragraph{Checkpoint delta.}
For each matched floating-point tensor with identical shape in the source and trained checkpoints, both tensors are first cast to bfloat16 and promoted to float32 before subtraction:
\begin{equation}
    \Delta W = W_{\mathrm{trained}} - W_{\mathrm{src}}.
\end{equation}
Frobenius norms and coordinate-energy sums use float64.
Update-matrix SVDs and source-coordinate mask construction use float32; source and trained singular-value spectra are computed directly from the two BF16-represented endpoints in float64.
BF16 is a standard checkpoint release precision: R1-Qwen, JustRL, MiniCPM5, and the Qwen source models studied here are released in BF16.
We use this common representation for all pairs, including checkpoints stored in FP32 such as DeepScaleR.
The metrics describe net changes between BF16-represented endpoints.
BF16 spacing increases with weight magnitude, so rounding contributes to which coordinate changes remain visible.

\paragraph{Aggregate relative update norm.}
For a collection \(B\) of matched tensors, such as all analyzed tensors or all tensors assigned to one layer--module group, the \textbf{Frobenius-aggregated relative update norm} is
\begin{equation}
    r(B)=
    \frac{\sqrt{\sum_{W\in B}\|\Delta W\|_F^2}}
         {\sqrt{\sum_{W\in B}\|W_{\mathrm{src}}\|_F^2}}.
\end{equation}
The metric measures parameter displacement relative to source-parameter energy.
Table~\ref{tab:global-delta} reports this quantity with \(B\) equal to all matched tensors.

\paragraph{Coordinate sparsity.}
We distinguish \textbf{exact-zero sparsity} from \textbf{visible sparsity}.
For a collection \(B\) of matched tensors, they are
\begin{equation}
\begin{aligned}
    s_0(B)
    &=
    \frac{\sum_{W\in B}|\{i:\Delta W_i=0\}|}
         {\sum_{W\in B}|W|},
    \\
    s_\epsilon^{\mathrm{vis}}(B)
    &=
    \frac{\sum_{W\in B}|\{i:|\Delta W_i|\leq\epsilon\}|}
         {\sum_{W\in B}|W|}.
\end{aligned}
\end{equation}
Exact-zero sparsity records coordinates that are identical in the BF16-represented endpoints.
The visible definition is equivalent to \(\mathrm{isclose}(\Delta W,0,\mathrm{atol}=\epsilon)\) because the comparison value is zero and the relative-tolerance term vanishes.
The main paper uses \(\epsilon=10^{-5}\); Table~\ref{tab:app-global-coordinate} additionally reports \(\epsilon\in\{10^{-6},10^{-4}\}\) and exact-zero sparsity.
All checkpoint- and group-level sparsities pool coordinate counts across tensors.

\paragraph{Relative-threshold sparsity.}
An absolute visibility threshold has the same numerical value for every tensor.
To account for differences in source-weight scale, \textbf{relative-threshold sparsity} instead scales the cutoff by each source tensor's RMS:
\begin{equation}
\begin{aligned}
    t_\tau(W)
    &=
    \tau\,\mathrm{RMS}(W_{\mathrm{src}}),
    \\
    s_\tau^{\mathrm{rel}}(B)
    &=
    \frac{
    \sum_{W\in B}
    |\{i:|\Delta W_i|<t_\tau(W)\}|}
    {\sum_{W\in B}|W|},
    \\
    \mathrm{RMS}(W_{\mathrm{src}})
    &=
    \sqrt{\frac{1}{|W|}\sum_i W_{\mathrm{src},i}^2}.
\end{aligned}
\end{equation}
Table~\ref{tab:app-global-energy} reports \(\tau=10^{-3}\).
As with visible sparsity, this aggregation weights each tensor by its number of coordinates.

\paragraph{Top-\(p\) coordinate energy ratio.}
This metric measures how much update energy is carried by the largest-magnitude coordinates.
For tensor \(W\), let \(n_W=|W|\), \(k_p=\max(1,\lceil pn_W\rceil)\), and let \(\mathrm{Top}_{k_p}(|\Delta W|)\) denote the indices of its \(k_p\) largest-magnitude update coordinates.
For a collection \(B\) with nonzero total update energy, we select the top \(p\) fraction separately within each tensor and pool their squared update energy:
\begin{equation}
    c_p(B)=
    \frac{
    \sum_{W\in B}
    \sum_{i\in\mathrm{Top}_{k_p}(|\Delta W|)}
    \Delta W_i^2}
    {\sum_{W\in B}\sum_i\Delta W_i^2}.
\end{equation}
Higher \(c_p(B)\) means that a smaller coordinate subset carries a larger fraction of the update energy.
Table~\ref{tab:app-global-energy} reports \(p\in\{0.001,0.01,0.05\}\); coordinates are ranked within each tensor.

\paragraph{Spectral metrics.}
For each 2D update matrix, let \(m=\min(d_{\mathrm{in}},d_{\mathrm{out}})\) and write
\begin{equation}
    \Delta W = U_\Delta \Sigma_\Delta V_\Delta^\top,
    \qquad
    \Sigma_\Delta=\mathrm{diag}(\sigma_1,\ldots,\sigma_m),
    \qquad
    \sigma_1\geq\cdots\geq\sigma_m\geq0.
\end{equation}
The reported top-\(k\), stable-rank, and normalized-rank values are unweighted medians over matrices with a complete exact singular spectrum.

\textbf{Top-\(k\) spectral energy} is the fraction of update energy captured by the leading \(k\) singular components:
\begin{equation}
    e_k(W)=\frac{\sum_{i=1}^{k}\sigma_i^2}{\sum_{i=1}^{m}\sigma_i^2}.
\end{equation}
High \(e_k\) means that a rank-\(k\) approximation captures a large fraction of the update's Frobenius energy.
Table~\ref{tab:app-spectral} reports \(k\in\{1,8,16\}\).

\textbf{Stable rank} measures how broadly the update energy is distributed across singular directions:
\begin{equation}
    \mathrm{srank}(\Delta W)=\frac{\|\Delta W\|_F^2}{\|\Delta W\|_2^2}
    =\frac{\sum_i\sigma_i^2}{\sigma_1^2}.
\end{equation}
It is close to \(1\) when the top singular value dominates and increases as energy spreads across more singular directions.

\textbf{Normalized numerical rank} is the fraction of singular values above a relative tolerance:
\begin{equation}
    \overline{\mathrm{rank}}_\tau(\Delta W)
    =\frac{|\{i:\sigma_i>\tau\sigma_1\}|}{m}.
\end{equation}
Lower values indicate a more sharply truncated spectrum at that tolerance.
The paper reports the median normalized rank at \(\tau=10^{-5}\).
Higher top-\(k\) energy and lower stable rank jointly indicate that more Frobenius energy is carried by the leading singular components, while normalized numerical rank records whether the spectrum is truncated at a fixed relative tolerance after accounting for matrix dimension.
An update can be numerically full-rank while concentrating much of its energy in a small number of singular directions.

\paragraph{Spectral drift from the source.}
For each eligible 2D matrix, \textbf{source-normalized spectral drift} measures the change in its singular values:
\begin{equation}
    d_{\mathrm{spec}}(W)=
    \frac{\|\sigma(W_{\mathrm{trained}})
    -\sigma(W_{\mathrm{src}})\|_2}
    {\|\sigma(W_{\mathrm{src}})\|_2}.
\end{equation}
This ratio is normalized by the source spectrum's Euclidean norm, not by the update norm.
It records changes in singular values and does not by itself measure singular-vector rotation.
The paper reports the unweighted median across matrices with exact source and trained spectra.

\paragraph{Source-coordinate mask overlap.}
\textbf{Principal-mask overlap} and \textbf{low-magnitude-mask overlap} express source geometry at individual parameter locations.
For a source SVD \(W_{\mathrm{src}}=U\Sigma V^\top\), let \(k=\max(1,\lfloor \rho\min(d_{\mathrm{in}},d_{\mathrm{out}})\rfloor)\).
The principal mask \(M_{\mathrm{prin}}^\rho\) contains the top \(\rho\) fraction of coordinates by magnitude in the rank-\(k\) source reconstruction \(U_k\Sigma_kV_k^\top\), while \(M_{\mathrm{low}}^\rho\) contains the bottom \(\rho\) fraction by \(|W_{\mathrm{src},i}|\).
Each mask contains \(\lceil\rho|W|\rceil\) coordinates.
Using the same threshold as the main sparsity analysis, the visible update support is
\(A_{10^{-5}}=\{i:|\Delta W_i|>10^{-5}\}\).
Its one-sided overlap with a source mask \(M\) is
\begin{equation}
    o(M)=\frac{|A_{10^{-5}}\cap M|}{|A_{10^{-5}}|}.
\end{equation}
Overlap is computed separately for each eligible 2D matrix, and Table~\ref{tab:global-delta} reports the unweighted median across matrices.
Under an independent update support, the expected overlap equals the realized density of \(M\), approximately \(\rho\); values below this baseline indicate avoidance, and values above it indicate enrichment.
Low principal-mask overlap means that visible updates avoid coordinates emphasized by the source's leading reconstruction, while high low-magnitude-mask overlap means that visible updates preferentially occupy coordinates where the source weight was small in magnitude.
The main table uses \(\rho=0.1\); Table~\ref{tab:app-geometry-overlap} additionally reports \(\rho\in\{0.01,0.05,0.2\}\).

\subsection{Update-support overlap metrics}
\label{app:overlap-metrics}

Table~\ref{tab:opd-rlvr-overlap} compares two visible update supports.
For checkpoint deltas \(\Delta^{(A)}\) and \(\Delta^{(B)}\), we define
\begin{equation}
    A=\{i:|\Delta^{(A)}_i|>10^{-5}\},
    \quad
    B=\{i:|\Delta^{(B)}_i|>10^{-5}\}.
\end{equation}
The directional overlap scores are
\begin{equation}
    o_{A\rightarrow B}=\frac{|A\cap B|}{|A|},
    \quad
    o_{B\rightarrow A}=\frac{|A\cap B|}{|B|}.
\end{equation}
These are asymmetric because each score normalizes by a different subnetwork size.
For example, if \(A\) is much smaller than \(B\), then most of \(A\) can overlap \(B\) even when only a modest fraction of \(B\) overlaps \(A\).

Let \(p_A=|A|/N\) and \(p_B=|B|/N\), where \(N\) is the number of matched coordinates.
Under independent supports, the directional baselines are
\begin{equation}
    \mathbb{E}_{\mathrm{ind}}[o_{A\rightarrow B}]=p_B,
    \qquad
    \mathbb{E}_{\mathrm{ind}}[o_{B\rightarrow A}]=p_A.
\end{equation}
Union sparsity is
\begin{equation}
    s_{\mathrm{union}}=1-\frac{|A\cup B|}{N},
\end{equation}
the fraction of coordinates updated by neither run.
Union sparsity describes the size of the union support and changes directly with the two marginal densities; it is therefore not itself a measure of overlap strength.

\subsection{Additional \textsc{OPD} update statistics}
\label{app:additional-static}

Tables~\ref{tab:app-global-coordinate}--\ref{tab:app-geometry-overlap} report additional statistics for the 13 checkpoint pairs summarized in the main table.
Gray rows are the same offline \textsc{SFT}/\textsc{SeqKD} and \textsc{RLVR} references used in the main paper.
The additional OpenThoughts3 run is analyzed in the support-overlap comparison (Table~\ref{tab:opd-rlvr-overlap}) and the training-trajectory analysis (Appendix~\ref{app:training-trajectory}).
All percentages are computed from final checkpoint deltas.
For R1-Qwen with the DeepScaleR teacher, exact-zero sparsity is \(82.40\%\) for \textsc{OPD} versus \(26.55\%\) for \textsc{SFT}; using \(10^{-3}\) times each source tensor's RMS gives \(89.51\%\) versus \(27.50\%\).
The separation thus persists across the absolute and relative thresholds reported here.
Exact-zero and visible sparsity use float32 differences of BF16-represented checkpoint weights, while relative-threshold sparsity uses the source-tensor RMS as its scale.
Spectral and source-geometry values are unweighted medians over eligible 2D matrices; source-mask overlap consistently uses the \(10^{-5}\) visible update support.

\begin{table*}[t]
\centering
\small
\setlength{\tabcolsep}{3.8pt}
\begin{tabular}{@{}lrrrrr@{}}
\toprule
Pair & Rel. norm & Exact zero & Sparsity@1e-6 & Sparsity@1e-5 & Sparsity@1e-4 \\
 & (\%) & (\%) & (\%) & (\%) & (\%) \\
\midrule
\textcolor{distillgray}{Qwen-Math Distill} & \textcolor{distillgray}{11.9363} & \textcolor{distillgray}{3.06} & \textcolor{distillgray}{3.06} & \textcolor{distillgray}{3.06} & \textcolor{distillgray}{3.52} \\
\textcolor{distillgray}{R1-Qwen SFT (DeepScaleR)} & \textcolor{distillgray}{0.8280} & \textcolor{distillgray}{26.55} & \textcolor{distillgray}{26.55} & \textcolor{distillgray}{26.63} & \textcolor{distillgray}{31.62} \\
\textcolor{rlvrgray}{DeepScaleR RLVR} & \textcolor{rlvrgray}{0.0725} & \textcolor{rlvrgray}{77.36} & \textcolor{rlvrgray}{77.37} & \textcolor{rlvrgray}{77.89} & \textcolor{rlvrgray}{94.21} \\
R1-Qwen OPD & 0.0488 & 82.40 & 82.41 & 83.04 & 97.91 \\
\textcolor{distillgray}{R1-Qwen SFT (JustRL)} & \textcolor{distillgray}{0.7188} & \textcolor{distillgray}{28.93} & \textcolor{distillgray}{28.94} & \textcolor{distillgray}{29.02} & \textcolor{distillgray}{34.82} \\
\textcolor{rlvrgray}{JustRL RLVR} & \textcolor{rlvrgray}{0.4973} & \textcolor{rlvrgray}{34.38} & \textcolor{rlvrgray}{34.39} & \textcolor{rlvrgray}{34.68} & \textcolor{rlvrgray}{49.16} \\
R1-Qwen OPD (JustRL) & 0.1165 & 69.31 & 69.34 & 69.82 & 85.12 \\
MiniCPM OPD & 0.0689 & 80.46 & 80.49 & 81.65 & 98.91 \\
\textcolor{rlvrgray}{Qwen2.5-VL RLVR} & \textcolor{rlvrgray}{0.0387} & \textcolor{rlvrgray}{84.65} & \textcolor{rlvrgray}{84.90} & \textcolor{rlvrgray}{87.99} & \textcolor{rlvrgray}{99.99} \\
Qwen2.5-VL OPD & 0.1421 & 64.17 & 64.47 & 66.80 & 94.82 \\
Qwen3 OPD & 0.0445 & 77.94 & 77.96 & 79.03 & 97.78 \\
Qwen3 OPSD & 0.1732 & 62.23 & 62.34 & 63.74 & 93.17 \\
TCOD-B2F OPD & 0.0633 & 77.71 & 77.86 & 81.38 & 99.98 \\
\bottomrule
\end{tabular}
\caption{Additional global norm and coordinate-sparsity statistics. Sparsity@\( \epsilon \) uses \(\mathrm{isclose}(\Delta W,0,\mathrm{atol}=\epsilon)\).}
\label{tab:app-global-coordinate}
\end{table*}

\begin{table*}[t]
\centering
\small
\setlength{\tabcolsep}{5.5pt}
\begin{tabular}{@{}lrrrr@{}}
\toprule
Pair & Rel.-thr. sparsity & Top 0.1\% coord. & Top 1\% coord. & Top 5\% coord. \\
 & (\%) & energy (\%) & energy (\%) & energy (\%) \\
\midrule
\textcolor{distillgray}{Qwen-Math Distill} & \textcolor{distillgray}{3.13} & \textcolor{distillgray}{1.85} & \textcolor{distillgray}{10.18} & \textcolor{distillgray}{30.86} \\
\textcolor{distillgray}{R1-Qwen SFT (DeepScaleR)} & \textcolor{distillgray}{27.50} & \textcolor{distillgray}{1.94} & \textcolor{distillgray}{11.56} & \textcolor{distillgray}{35.09} \\
\textcolor{rlvrgray}{DeepScaleR RLVR} & \textcolor{rlvrgray}{83.37} & \textcolor{rlvrgray}{3.77} & \textcolor{rlvrgray}{21.18} & \textcolor{rlvrgray}{53.00} \\
R1-Qwen OPD & 89.51 & 3.78 & 19.80 & 58.44 \\
\textcolor{distillgray}{R1-Qwen SFT (JustRL)} & \textcolor{distillgray}{30.05} & \textcolor{distillgray}{1.98} & \textcolor{distillgray}{11.76} & \textcolor{distillgray}{36.51} \\
\textcolor{rlvrgray}{JustRL RLVR} & \textcolor{rlvrgray}{37.33} & \textcolor{rlvrgray}{8.39} & \textcolor{rlvrgray}{27.12} & \textcolor{rlvrgray}{55.98} \\
R1-Qwen OPD (JustRL) & 73.81 & 1.75 & 12.57 & 43.52 \\
MiniCPM OPD & 89.05 & 2.80 & 22.93 & 58.05 \\
\textcolor{rlvrgray}{Qwen2.5-VL RLVR} & \textcolor{rlvrgray}{93.21} & \textcolor{rlvrgray}{4.03} & \textcolor{rlvrgray}{32.06} & \textcolor{rlvrgray}{71.54} \\
Qwen2.5-VL OPD & 70.57 & 3.09 & 13.45 & 45.86 \\
Qwen3 OPD & 87.90 & 3.33 & 19.09 & 59.03 \\
Qwen3 OPSD & 66.83 & 1.89 & 11.23 & 39.48 \\
TCOD-B2F OPD & 86.67 & 2.67 & 17.17 & 57.46 \\
\bottomrule
\end{tabular}
\caption{Relative-threshold sparsity and coordinate-energy statistics. Rel.-thr. sparsity uses \(10^{-3}\cdot\mathrm{RMS}(W_{\mathrm{src}})\). Coordinate-energy values select the largest update coordinates within each tensor and then pool their squared update energy.}
\label{tab:app-global-energy}
\end{table*}

\begin{table*}[t]
\centering
\scriptsize
\begin{tabular*}{\textwidth}{@{\extracolsep{\fill}}lrrrrrr@{}}
\toprule
Pair & Top-1 SVD & Top-8 SVD & Top-16 SVD & Stable rank & Rank@1e-5 & Spectral drift \\
 & energy (\%) & energy (\%) & energy (\%) & & (\%) & \\
\midrule
\textcolor{distillgray}{Qwen-Math Distill} & \textcolor{distillgray}{1.21} & \textcolor{distillgray}{4.99} & \textcolor{distillgray}{8.57} & \textcolor{distillgray}{82.75} & \textcolor{distillgray}{100} & \textcolor{distillgray}{\(2.958{\times}10^{-2}\)} \\
\textcolor{distillgray}{R1-Qwen SFT (DeepScaleR)} & \textcolor{distillgray}{2.15} & \textcolor{distillgray}{9.64} & \textcolor{distillgray}{15.34} & \textcolor{distillgray}{46.48} & \textcolor{distillgray}{100} & \textcolor{distillgray}{\(1.782{\times}10^{-4}\)} \\
\textcolor{rlvrgray}{DeepScaleR RLVR} & \textcolor{rlvrgray}{6.22} & \textcolor{rlvrgray}{16.13} & \textcolor{rlvrgray}{19.62} & \textcolor{rlvrgray}{16.07} & \textcolor{rlvrgray}{100} & \textcolor{rlvrgray}{\(2.023{\times}10^{-5}\)} \\
R1-Qwen OPD & 10.92 & 24.08 & 28.69 & 9.15 & 100 & \(1.387{\times}10^{-5}\) \\
\textcolor{distillgray}{R1-Qwen SFT (JustRL)} & \textcolor{distillgray}{3.62} & \textcolor{distillgray}{12.28} & \textcolor{distillgray}{18.50} & \textcolor{distillgray}{27.60} & \textcolor{distillgray}{100} & \textcolor{distillgray}{\(1.422{\times}10^{-4}\)} \\
\textcolor{rlvrgray}{JustRL RLVR} & \textcolor{rlvrgray}{8.09} & \textcolor{rlvrgray}{20.76} & \textcolor{rlvrgray}{26.94} & \textcolor{rlvrgray}{12.35} & \textcolor{rlvrgray}{100} & \textcolor{rlvrgray}{\(2.043{\times}10^{-3}\)} \\
R1-Qwen OPD (JustRL) & 14.05 & 36.14 & 42.87 & 7.12 & 100 & \(3.612{\times}10^{-5}\) \\
MiniCPM OPD & 8.11 & 21.50 & 26.81 & 12.32 & 100 & \(1.184{\times}10^{-5}\) \\
\textcolor{rlvrgray}{Qwen2.5-VL RLVR} & \textcolor{rlvrgray}{5.78} & \textcolor{rlvrgray}{16.21} & \textcolor{rlvrgray}{20.46} & \textcolor{rlvrgray}{17.30} & \textcolor{rlvrgray}{100} & \textcolor{rlvrgray}{\(9.929{\times}10^{-6}\)} \\
Qwen2.5-VL OPD & 8.14 & 23.29 & 30.52 & 12.28 & 100 & \(3.194{\times}10^{-5}\) \\
Qwen3 OPD & 4.92 & 14.27 & 19.69 & 20.30 & 100 & \(1.824{\times}10^{-5}\) \\
Qwen3 OPSD & 11.74 & 24.30 & 29.49 & 8.52 & 100 & \(2.823{\times}10^{-5}\) \\
TCOD-B2F OPD & 7.42 & 18.44 & 23.64 & 13.46 & 100 & \(1.202{\times}10^{-5}\) \\
\bottomrule
\end{tabular*}
\caption{Additional spectral statistics. Top-\(k\) SVD energy, stable rank, normalized numerical rank, and spectral drift are unweighted medians over eligible 2D matrices with exact computations.}
\label{tab:app-spectral}
\end{table*}

\FloatBarrier

\begin{table}[!ht]
\centering
\scriptsize
\begin{tabular*}{\linewidth}{@{\extracolsep{\fill}}lrrrrrrrr@{}}
\toprule
& \multicolumn{4}{c}{Source-principal overlap (\%)} & \multicolumn{4}{c}{Low-magnitude overlap (\%)} \\
\cmidrule(lr){2-5}\cmidrule(lr){6-9}
Pair & 1\% & 5\% & 10\% & 20\% & 1\% & 5\% & 10\% & 20\% \\
\midrule
\textcolor{distillgray}{Qwen-Math Distill} & \textcolor{distillgray}{1.00} & \textcolor{distillgray}{4.98} & \textcolor{distillgray}{9.95} & \textcolor{distillgray}{19.90} & \textcolor{distillgray}{1.01} & \textcolor{distillgray}{5.05} & \textcolor{distillgray}{10.09} & \textcolor{distillgray}{20.18} \\
\textcolor{distillgray}{R1-Qwen SFT (DeepScaleR)} & \textcolor{distillgray}{0.87} & \textcolor{distillgray}{4.39} & \textcolor{distillgray}{8.76} & \textcolor{distillgray}{17.64} & \textcolor{distillgray}{1.18} & \textcolor{distillgray}{5.88} & \textcolor{distillgray}{11.76} & \textcolor{distillgray}{23.29} \\
\textcolor{rlvrgray}{DeepScaleR RLVR} & \textcolor{rlvrgray}{0.58} & \textcolor{rlvrgray}{2.82} & \textcolor{rlvrgray}{5.31} & \textcolor{rlvrgray}{10.18} & \textcolor{rlvrgray}{2.95} & \textcolor{rlvrgray}{14.74} & \textcolor{rlvrgray}{29.73} & \textcolor{rlvrgray}{54.13} \\
R1-Qwen OPD & 0.59 & 2.75 & 4.98 & 9.20 & 3.64 & 18.20 & 36.68 & 65.10 \\
\textcolor{distillgray}{R1-Qwen SFT (JustRL)} & \textcolor{distillgray}{0.85} & \textcolor{distillgray}{4.28} & \textcolor{distillgray}{8.49} & \textcolor{distillgray}{17.17} & \textcolor{distillgray}{1.22} & \textcolor{distillgray}{6.11} & \textcolor{distillgray}{12.23} & \textcolor{distillgray}{24.15} \\
\textcolor{rlvrgray}{JustRL RLVR} & \textcolor{rlvrgray}{0.86} & \textcolor{rlvrgray}{4.24} & \textcolor{rlvrgray}{8.35} & \textcolor{rlvrgray}{16.59} & \textcolor{rlvrgray}{1.44} & \textcolor{rlvrgray}{7.20} & \textcolor{rlvrgray}{14.44} & \textcolor{rlvrgray}{28.03} \\
R1-Qwen OPD (JustRL) & 0.66 & 3.05 & 5.82 & 11.17 & 2.25 & 11.23 & 22.55 & 42.64 \\
MiniCPM OPD & 0.67 & 2.93 & 5.41 & 9.48 & 3.63 & 18.01 & 36.95 & 65.78 \\
\textcolor{rlvrgray}{Qwen2.5-VL RLVR} & \textcolor{rlvrgray}{0.60} & \textcolor{rlvrgray}{2.59} & \textcolor{rlvrgray}{4.70} & \textcolor{rlvrgray}{7.93} & \textcolor{rlvrgray}{3.75} & \textcolor{rlvrgray}{18.35} & \textcolor{rlvrgray}{38.30} & \textcolor{rlvrgray}{70.94} \\
Qwen2.5-VL OPD & 0.64 & 2.91 & 5.34 & 9.90 & 2.21 & 11.01 & 22.05 & 43.80 \\
Qwen3 OPD & 0.54 & 2.44 & 4.46 & 8.05 & 3.13 & 15.44 & 32.06 & 59.20 \\
Qwen3 OPSD & 0.63 & 3.03 & 5.70 & 10.66 & 2.06 & 10.24 & 20.66 & 40.59 \\
TCOD-B2F OPD & 0.55 & 2.75 & 5.29 & 9.84 & 2.58 & 12.95 & 24.75 & 53.84 \\
\bottomrule
\end{tabular*}
\caption{Source-relative coordinate-mask overlap. Values are unweighted medians over eligible 2D matrices of the fraction of \(10^{-5}\)-visible update coordinates falling in source-principal or low-magnitude coordinate masks at each mask density.}
\label{tab:app-geometry-overlap}
\end{table}

\FloatBarrier

\raggedbottom

\subsection{Component-wise updates in the VLM}
\label{app:vlm-components}

For the Qwen2.5-VL \textsc{OPD} checkpoint, we separate the vision encoder, multimodal merger, and language backbone.
All three components are trained, totaling 3.755 billion independent parameters.
The merger comprises \texttt{visual.merger.*}; the vision encoder contains the remaining visual parameters, and the language backbone includes the shared token embedding and output head, counted once.

\begin{table}[ht]
\centering
\small
\setlength{\tabcolsep}{3pt}
\begin{tabular*}{\linewidth}{@{\extracolsep{\fill}}lrrrrrr@{}}
\toprule
Component & Params. & Rel. norm & Sparsity & Energy share & Stable rank & Top-16 energy \\
 & (M) & (\%) & (\%) & (\%) & & (\%) \\
\midrule
Vision encoder & 631.98 & 0.1330 & 65.31 & 11.54 & 9.51 & 31.44 \\
Merger & 36.71 & 0.1568 & 64.97 & 1.15 & 9.06 & 34.20 \\
Language backbone & 3085.94 & 0.1433 & 67.13 & 87.32 & 14.13 & 29.48 \\
\bottomrule
\end{tabular*}
\caption{Component-wise geometry of Qwen2.5-VL \textsc{OPD}. Relative norm and sparsity pool parameters within each component, with \(\epsilon=10^{-5}\) at BF16 precision. Energy share is the component's fraction of total squared update norm. Spectral statistics are unweighted medians over nonzero 2D updates.}
\label{tab:vlm-components}
\end{table}

All three components exhibit similar visible sparsity (approximately \(65\%\)--\(67\%\)).
The language backbone accounts for \(87.32\%\) of update energy and \(82.19\%\) of parameters, while the merger has the largest relative update norm.
The vision encoder and merger have lower median stable rank and higher Top-16 energy than the language backbone.
The spectral summaries cover 160 vision matrices, both merger matrices (\(5120\times5120\) and \(2048\times5120\)), and 253 language matrices, all with exact SVDs; the 5D patch embedding and 1D parameters contribute to norm and sparsity but not spectral statistics.

\FloatBarrier

\subsection{Evolution of sparsity and stable rank along training}
\label{app:training-trajectory}

\begin{figure*}[t]
\centering
\begin{minipage}[t]{0.49\textwidth}
\centering
\includegraphics[width=\linewidth]{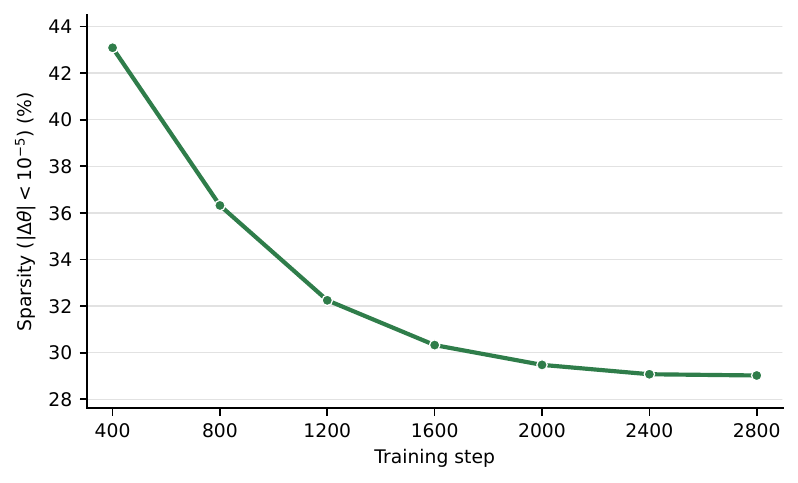}\par\vspace{-2pt}
{\small\textbf{(a)} R1-Qwen \textsc{SFT} (JustRL): sparsity}
\end{minipage}
\hfill
\begin{minipage}[t]{0.49\textwidth}
\centering
\includegraphics[width=\linewidth]{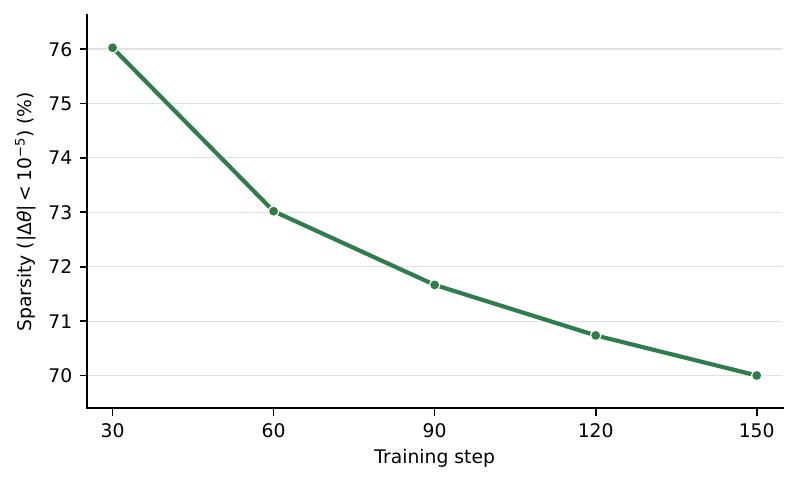}\par\vspace{-2pt}
{\small\textbf{(b)} \textsc{OPD} (JustRL / OpenThoughts3): sparsity}
\end{minipage}
\par\vspace{4pt}
\begin{minipage}[t]{0.49\textwidth}
\centering
\includegraphics[width=\linewidth]{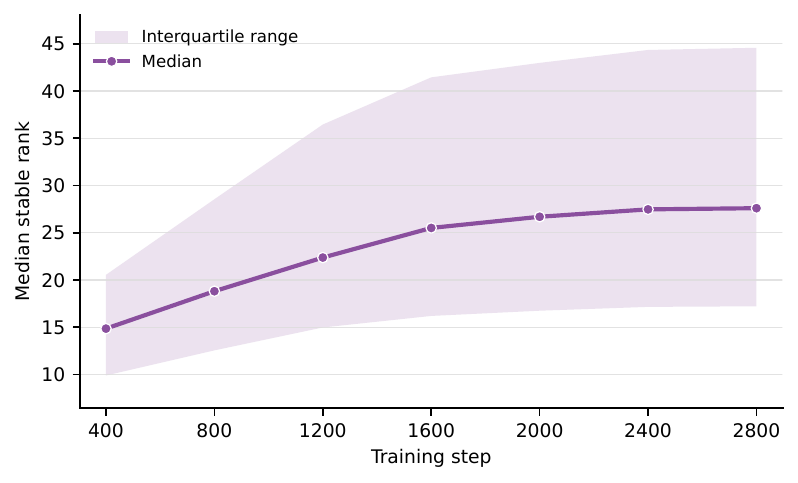}\par\vspace{-2pt}
{\small\textbf{(c)} R1-Qwen \textsc{SFT} (JustRL): stable rank}
\end{minipage}
\hfill
\begin{minipage}[t]{0.49\textwidth}
\centering
\includegraphics[width=\linewidth]{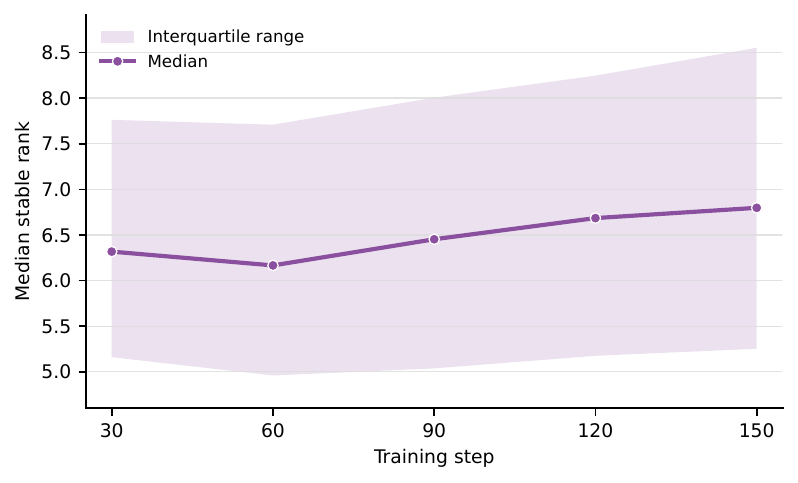}\par\vspace{-2pt}
{\small\textbf{(d)} \textsc{OPD} (JustRL / OpenThoughts3): stable rank}
\end{minipage}
\caption{Training-trajectory evolution of checkpoint-delta sparsity and stable rank. Top: global \(10^{-5}\)-visible sparsity. Bottom: median stable rank (line) and interquartile range (shading) over eligible two-dimensional update matrices. Both runs use the JustRL teacher; SFT uses DAPO-Math-17K and OPD uses OpenThoughts3. Each delta is measured relative to the common R1-Qwen source. The trajectories describe runs with different datasets and training schedules; their step axes are not matched for data exposure or compute.}
\label{fig:training-trajectory-sparsity-rank}
\end{figure*}

To complement the final-checkpoint statistics, Figure~\ref{fig:training-trajectory-sparsity-rank} tracks deltas from the same DeepSeek-R1-Distill-Qwen-1.5B source to intermediate checkpoints of two representative runs.
Both runs use the JustRL teacher: we analyze seven R1-Qwen \textsc{SFT} checkpoints trained on DAPO-Math-17K at steps \(\{400,800,\ldots,2800\}\) and five R1-Qwen \textsc{OPD} checkpoints trained on OpenThoughts3 at steps \(\{30,60,\ldots,150\}\).
At each checkpoint, visible sparsity is the fraction of all 1.777 billion matched floating-point parameters satisfying \(|\Delta W|\leq10^{-5}\), while stable rank is computed separately for each of 198 eligible two-dimensional update matrices.
The stable-rank curves report the median and interquartile range across matrices, using the definitions in Section~\ref{app:delta-metrics}.

Both runs become less sparse as training proceeds, but their spectral trajectories differ over the observed checkpoints.
For R1-Qwen \textsc{SFT} (JustRL), sparsity decreases from \(43.09\%\) to \(29.02\%\) while median stable rank rises from \(14.86\) to \(27.60\), indicating that the accumulated update spreads across both more coordinates and more singular directions.
For R1-Qwen \textsc{OPD} on OpenThoughts3 with the JustRL teacher, sparsity decreases from \(76.02\%\) to \(70.00\%\), whereas median stable rank remains between \(6.17\) and \(6.80\).
Thus, the visible \textsc{OPD} support expands without comparable spectral broadening, consistent with the final-checkpoint pattern of sparse, spectrally concentrated \textsc{OPD} updates.

\subsection{Top-1\% SVD reconstruction protocol}
\label{app:top1-svd-reconstruction}

This section details the functional spectral intervention in Section~\ref{subsec:functional-top-rank}.
We use DeepSeek-R1-Distill-Qwen-1.5B as the common source and compare the DeepScaleR-teacher \textsc{OPD} and fixed-data \textsc{SFT} checkpoints.
For each of the 196 two-dimensional weight matrices in the self-attention and MLP modules, we form \(\Delta W_p=W_{p,\mathrm{trained}}-W_{p,\mathrm{src}}\) in float32 and compute an exact SVD with full matrices disabled.
We retain \(q_p=\lceil0.01\min(d_{\mathrm{out}},d_{\mathrm{in}})\rceil\) leading components: \(q_p=16\) for the query, output, and MLP projections, and \(q_p=3\) for the key and value projections in R1-Qwen.
The truncated update is rescaled independently for each matrix:
\begin{equation}
\begin{aligned}
    \widehat{\Delta W}_{p,1\%}
    &=
    \frac{\|\Delta W_p\|_F}
         {\|\Delta W_{p,1\%}\|_F}
    \Delta W_{p,1\%},
    \\
    W_{p,\mathrm{recon}}
    &=
    W_{p,\mathrm{src}}+\widehat{\Delta W}_{p,1\%}.
\end{aligned}
\end{equation}
Embedding, language-head, bias, and normalization parameters are kept at their source values.
Per-matrix rescaling restores the original update norm after truncation, controlling update magnitude when evaluating the leading singular subspace.

We evaluate Base, Full, and reconstructed checkpoints on AIME24 and AIME25 with temperature \(0.7\), top-\(p=0.95\), 16 responses per problem, and a maximum generation length of 31,744 tokens.
The gain-recovery metric is computed benchmark-wise against the same Base checkpoint.

\FloatBarrier

\subsection{Benchmark-wise breakdown of the subnetwork intervention}
\label{app:benchmark-wise-functional}

\begin{figure}[ht]
\centering
\includegraphics[width=0.49\linewidth]{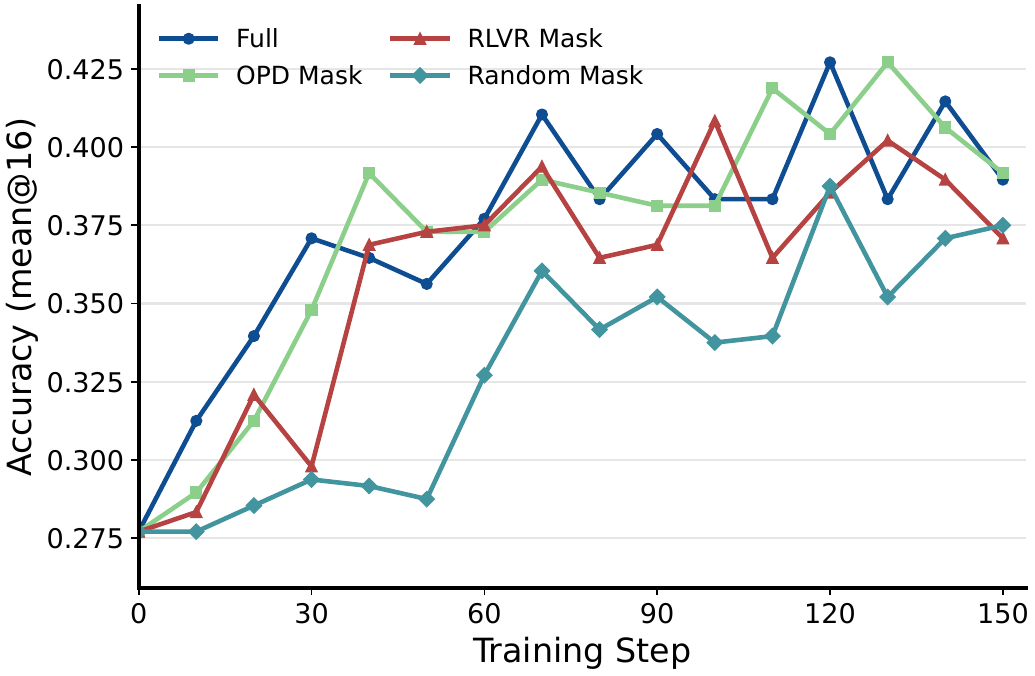}\hfill
\includegraphics[width=0.49\linewidth]{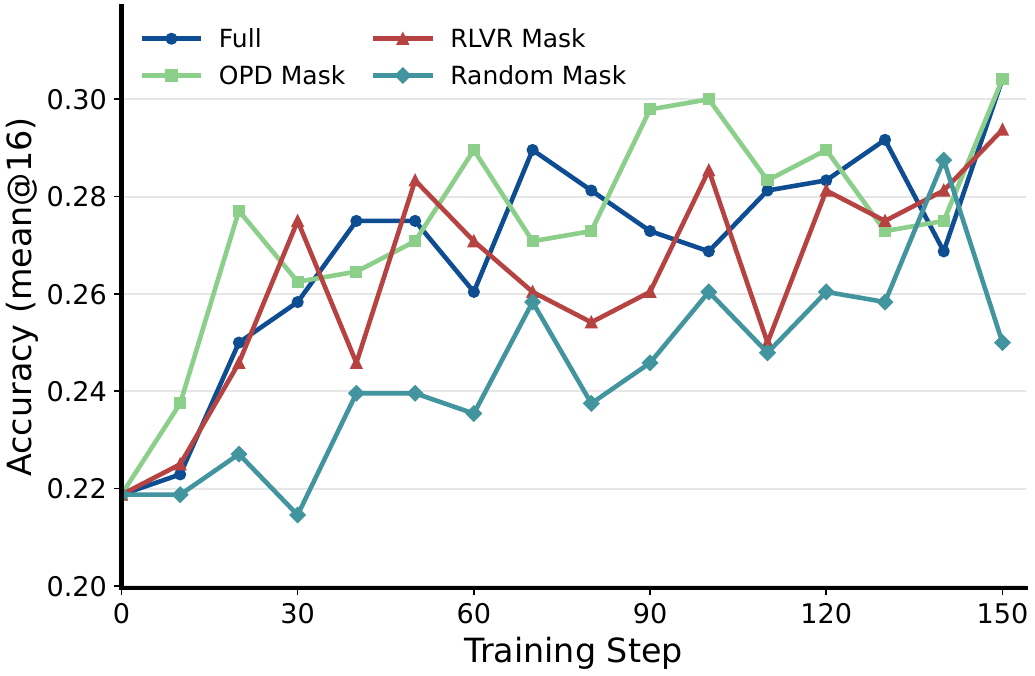}
\caption{Benchmark-wise breakdown of the R1-Qwen subnetwork-masked \textsc{OPD} experiment in Figure~\ref{fig:functional-average}. Left: AIME24. Right: AIME25. The \textsc{OPD} mask tracks the full run closely on both benchmarks, while the random mask remains weaker.}
\label{fig:functional-subnetwork-breakdown}
\end{figure}

The benchmark-wise curves support the same conclusion as the averaged curve in Figure~\ref{fig:functional-average}: the \textsc{OPD} mask closely tracks full training on both AIME24 and AIME25, while the random-mask run generally performs worse.

\end{document}